\documentclass[sigconf]{acmart}

\usepackage[textsize=tiny]{todonotes}
\usepackage[T1]{fontenc}
\usepackage{graphicx}
\usepackage{appendix}
\usepackage{comment}
\usepackage{enumitem}
\usepackage{tabularx}
\usepackage{booktabs}
\usepackage{fancyvrb} 
\usepackage{xcolor} 
\usepackage{tikz}
\usepackage{multirow}
\usepackage{longtable}
\usepackage{array}     
\usepackage{caption}   
\usepackage{placeins}
\usepackage{enumitem}
\usepackage{tikz}
\usetikzlibrary{shapes.geometric, arrows, positioning}
\tikzstyle{startstop} = [rectangle, rounded corners, minimum width=3cm, minimum height=1cm, text centered, draw=black, fill=cyan!30]
\tikzstyle{process} = [rectangle, minimum width=3cm, minimum height=1cm, text centered, draw=black, fill=orange!30]
\tikzstyle{miniprocess} = [rectangle, minimum width=2cm, minimum height=1cm, text centered, draw=black, fill=orange!30]
\tikzstyle{decision} = [diamond, minimum width=3cm, minimum height=1cm, text centered, draw=black, fill=red!30]
\tikzstyle{arrow} = [thick,->,>=stealth]
\usetikzlibrary{calc}
\usepackage{courier}
\usepackage{fontawesome5}
\usepackage{listings}
\usepackage{subcaption}
\usepackage{tcolorbox}  
\usepackage{tikz}
\usetikzlibrary{arrows.meta, positioning}
\usepackage{pgfgantt}
\usepackage{graphicx}

\lstdefinestyle{PythonStyle}{
    language=Python,
    basicstyle=\ttfamily\footnotesize,
    keywordstyle=\color{blue},
    commentstyle=\color{green},
    breaklines=true,
    frame=single
}

\lstdefinestyle{JSONStyle}{
    language=Java,
    basicstyle=\ttfamily\footnotesize,
    keywordstyle=\color{blue},
    breaklines=true,
    frame=single
}

\usepackage[utf8]{inputenc} 

\acmConference[ACM]{2025}{under review}{}

\begin{document}

\title{REALM-Bench: A Benchmark for Evaluating Multi-Agent Systems on Real-world, Dynamic Planning and Scheduling Tasks}

\author{Longling Geng}
\email{gll2027@stanford.edu}
\affiliation{%
  \institution{Stanford University}
  \city{Stanford}
  \state{CA}
  \country{USA}
}

\author{Edward Y. Chang}
\email{echang@cs.stanford.edu}
\affiliation{%
  \institution{Stanford University}
  \city{Stanford}
  \state{CA}
  \country{USA}
}

\renewcommand{\shortauthors}{}


\begin{abstract}
This benchmark suite provides a comprehensive evaluation framework for assessing both individual LLMs and multi-agent systems in Real-world planning and scheduling scenarios. The suite encompasses 14 designed planning and scheduling problems that progress from basic to highly complex, incorporating key aspects such as multi-agent coordination, inter-agent dependencies, and dynamic environmental disruptions. Each problem can be scaled along three dimensions: the number of parallel planning threads, the complexity of inter-dependencies, and the frequency of unexpected disruptions requiring Real-time adaptation. The benchmark includes 14 detailed problem specifications, 15 comparison methods including Random, LPT, SPT, STPT, MPSR, DRL-Liu, GP, GEP, LSO, SPT/TWKR, DRL-Chen, DRL-Zhang, 2+ evaluation metrics, and baseline implementations using 3+ LLMs including GPT-4o, Claude-3.7, DeepSeek-R1, and 4 contemporary frameworks including LangGraph, AutoGen, CrewAI, and Swarm, enabling rigorous testing of both single-agent and multi-agent planning capabilities. Through standardized evaluation criteria and scalable complexity, this benchmark aims to be opened to public, and drive progress in developing more adaptable, robust, and scalable AI planning systems for Real-world applications. \\
\textbf{Code:}{https://github.com/genglongling/REALM-Bench}\\
\textbf{Datasets:}{https://github.com/genglongling/M-APPLE-OS} \\
\end{abstract}



\maketitle

\section{Introduction}
Large language models (LLMs) today excel at single-turn interactions: answering questions, generating text, solving isolated reasoning problems, and providing explanations within individual conversations. However, real-world applications demand capabilities that current LLMs struggle with: multi-step planning over extended horizons, coordinated multi-agent collaboration, adaptive responses to dynamic environments, and sustained problem-solving that requires persistent state management and iterative refinement.

Recent advances in LLMs, including OpenAI's GPT-4o-Task~\cite{openai2024gpt4o}, DeepSeek's R1~\cite{deepseekai2025deepseekr1}, Anthropic's Claude 3.5 Sonnet~\cite{anthropic2024claude}, and Gemini~\cite{geminiteam2023gemini}, demonstrate impressive natural language capabilities but remain limited when faced with complex, multi-faceted problems that require orchestrating multiple agents, maintaining coherent plans across disruptions, ensuring transaction integrity across distributed operations, and adapting strategies based on evolving constraints and objectives.

Critical gaps persist in three key areas. First, \textit{temporal reasoning}: current systems struggle with long-horizon planning where early decisions impact future possibilities and constraints evolve dynamically. Second, \textit{collaborative intelligence}: while individual agents reason well, they fail at knowledge sharing, role specialization, and conflict resolution when working together. Third, \textit{resilience}: systems cannot gracefully handle disruptions, maintain consistency across distributed decision-making, or perform effective recovery operations when plans fail.

This recognition has driven the development of multi-agent frameworks including AutoGen~\cite{wu2024autogen}, CAMEL~\cite{li2023camel}, CrewAI~\cite{crewai2024}, LangGraph~\cite{langgraph2024}, DSPy~\cite{khattab2023dspy}, and XAgent~\cite{xia2023xagent}. However, these systems remain largely untested in standardized and challenging problems that mirror real-world complexity.

\paragraph{The evaluation gap is critical.} Existing AI benchmarks focus on perception, language understanding, or individual reasoning tasks—none systematically evaluate multi-agent coordination under realistic constraints. Without comprehensive benchmarks, researchers cannot compare approaches, identify bottlenecks, or validate improvements. Practitioners cannot confidently deploy systems in high-stakes domains where coordination failures have serious consequences.

We address this gap by introducing a comprehensive evaluation framework for multi-agent planning and coordination across 14 challenging problems spanning logistics, scheduling, resource allocation, and crisis response.

\subsection{The REALM Benchmark Suite}

REALM-Bench (REliable multi-Agent Logistics Management benchmark) is an extensible evaluation suite designed to test the planning, coordination, and adaptation abilities of LLMs and agentic systems under both nominal and disrupted conditions. A key distinguishing feature of REALM-Bench is its systematic focus on evaluating system reliability when facing unexpected events such as equipment failures, resource shortages, timing conflicts, and environmental changes that require dynamic replanning and recovery.

To ensure comprehensive coverage across problem domains, REALM-Bench incorporates two complementary benchmark categories. First, it includes ten \textbf{P}-series problems covering real-world planning scenarios that progress from static single-agent tasks to dynamic multi-agent coordination under severe disruptions. Second, REALM-Bench incorporates Job-Shop Scheduling Problems (JSSP)~\cite{DEMIRKOL1998137,destouet2023flexible} through four \textbf{J}-series problems, representing a canonical class of combinatorial optimization problems extensively studied by the operations research and manufacturing communities for over three decades.

The inclusion of JSSP provides well-defined mathematical foundations for evaluating planning optimality, constraint satisfaction, and computational efficiency. With the emerging natural language reasoning and multi-agent coordination capabilities of LLMs, it is timely to investigate whether these models can introduce adaptive intelligence into traditional optimization frameworks, particularly for disruptive, reactive, and proactive planning scenarios that exceed the scope of conventional algorithmic approaches.

\subsubsection{Task Difficulty Progression: Static to Dynamic, Simple to Complex}

REALM-Bench organizes 14 evaluation scenarios along a carefully designed difficulty progression, testing systems across five key dimensions: planning complexity, inter-agent dependencies, constraint satisfaction, dynamic disruptions, and reactive replanning requirements. These benchmark scenarios are labeled with \textbf{P} for general planning problems and \textbf{J} for JSSP variants.

\vspace{0.3em}
\noindent\textbf{Tier 1: Single-Agent Static Planning (Foundational)}
\begin{itemize}[leftmargin=1.2em, topsep=-.05em, parsep=-.05em]
    \item \textbf{P1: Campus Tour (CT-static)} -- Single-agent navigation with time windows and spatial constraints
    \item \textbf{J1: JSSP Basic (static)} -- Basic job-shop scheduling with fixed machine assignments
\end{itemize}

\vspace{0.3em}
\noindent\textbf{Tier 2: Multi-Agent Static Coordination (Moderate)}
\begin{itemize}[leftmargin=1.2em, topsep=-.05em, parsep=-.05em]
    \item \textbf{P2: Multi-Group Campus Tours (MCT-static)} -- Multiple coordinated tours with shared resource constraints
    \item \textbf{P3: Urban Ride-Sharing (URS-static)} -- Vehicle-passenger matching with capacity and timing constraints
    \item \textbf{P5: Wedding Reunion Logistics (WR-static)} -- Group coordination with shared transportation
\end{itemize}

\vspace{0.3em}
\noindent\textbf{Tier 3: Complex Static Dependencies (Challenging)}
\begin{itemize}[leftmargin=1.2em, topsep=-.05em, parsep=-.05em]
    \item \textbf{P6: Thanksgiving Dinner Planning (TD-static)} -- Multi-domain coordination with travel and preparation dependencies
    \item \textbf{P7: Disaster Relief Logistics (DL-static)} -- Resource allocation under uncertainty with priority constraints
    \item \textbf{J3: JSSP Large-scale (static)} -- Advanced job-shop scheduling with complex machine-job mappings
\end{itemize}

\vspace{0.3em}
\noindent\textbf{Tier 4: Dynamic Disruption Handling (Advanced)}
\begin{itemize}[leftmargin=1.2em, topsep=-.05em, parsep=-.05em]
    \item \textbf{P4: URS with Disruptions (URS-dynamic)} -- Real-time replanning with traffic delays and route changes
    \item \textbf{P8: Wedding Logistics with Disruptions (WR-dynamic)} -- Recovery from road closures and conflicts
    \item \textbf{P9: Thanksgiving with Disruptions (TD-dynamic)} -- Flight delays requiring coordinated recovery
    \item \textbf{J2, J4: JSSP with Disruptions (dynamic)} -- Machine breakdowns requiring real-time rescheduling
\end{itemize}

\vspace{0.3em}
\noindent\textbf{Tier 5: Large-Scale Multi-Domain Integration (Expert)}
\begin{itemize}[leftmargin=1.2em, topsep=-.05em, parsep=-.05em]
    \item \textbf{P10: Global Supply Chain (GSC-static/dynamic)} -- Enterprise-scale planning with procurement, infrastructure dependencies, and cost-risk optimization
\end{itemize}

\subsubsection{Task Scalability: Small to Large Instances}

Beyond difficulty progression, each problem supports \textit{scalability} along instance size dimensions—a separate axis from difficulty. For example, urban ride-sharing can scale from 3 vehicles and 10 passengers (small instance) to 50 vehicles and 500 passengers (large instance) while maintaining the same coordination complexity. Similarly, JSSP scenarios scale from 3×3 (jobs×machines) to 20×20 configurations.

Crucially, \textit{small instances are not necessarily easy}. A small-scale disaster relief scenario with complex dependencies can be more challenging than a large-scale ride-sharing problem with simple constraints. This scalability dimension enables systematic evaluation under increasing computational demands while allowing detailed failure analysis in manageable configurations, providing orthogonal stress-testing to the difficulty tiers.

\section{Related Benchmarks}

Planning benchmarks have evolved from testing basic STRIPS-style planning to evaluating increasingly sophisticated capabilities. The International Planning Competition (IPC) has driven planning benchmarks since 1998, using PDDL to specify domains like BlocksWorld, Logistics, and Rovers~\cite{PlanningCompetition2024}. While valuable for classical planning algorithms, these benchmarks focus on deterministic environments with complete information and lack dynamic disruptions common in real-world scenarios.

Recent benchmarks have expanded scope. The Process Planning Competition (PPC) addresses continuous processes and temporal constraints in manufacturing scenarios~\cite{ppc2020}, while the Dynamic Planning Competition introduces environmental changes during execution~\cite{dpc2022}. The Automated Negotiation Agents Competition (ANAC) incorporates planning elements within supply chain scenarios~\cite{anac2023}, though it focuses primarily on bilateral negotiations rather than comprehensive planning under uncertainty.

For LLM planning capabilities,  TimeBench~\cite{chu2023timebench} tests temporal reasoning and scheduling constraints, TaskBench~\cite{shen2023taskbench} evaluates practical task automation and step-by-step planning, and PlanBench~\cite{valmeekam2023planbench} provides end-to-end plan evaluation across static datasets. However, these often rely on oversimplified synthetic scenarios.

This landscape reveals key gaps in existing benchmarks:

\begin{enumerate}[leftmargin=1.5em, topsep=0pt, parsep=0pt, label=\arabic*.]
\item \textbf{Limited Dynamic Complexity:} Most benchmarks treat disruptions as isolated events rather than cascading, interdependent phenomena that create complex dependency networks requiring sophisticated coordination.

\item \textbf{Missing Transaction Properties:} No benchmarks evaluate distributed planning systems for transactional integrity, compensation mechanisms, rollback capabilities, or state consistency across multiple agents when plans fail or require modification.

\item \textbf{Short-Lived Episode Focus:} Existing benchmarks assume finite planning episodes with clear start/end points, ignoring long-lived workflows that persist across disruptions, require continuous adaptation, and maintain operational continuity indefinitely.

\item \textbf{Representation Limitations:} Simplified formats like PDDL cannot capture multifaceted real-world constraints, multi-objective optimization, or the rich semantic relationships between entities in complex domains.

\item \textbf{Narrow Scope Evaluation:} Benchmarks focus on specific subproblems (path planning, resource allocation) rather than end-to-end scenarios that combine planning, execution monitoring, failure recovery, and continuous replanning.
\end{enumerate}

REALM-Bench addresses these limitations by providing scenarios that combine complex dependencies, dynamic disruptions, and systematic scalability while remaining tractable for evaluation. This enables testing planning systems under conditions that better reflect real-world application challenges.
\section{Benchmark Problem Specifications}

We detail the 14 benchmark problems spanning five difficulty tiers from foundational single-agent planning to expert-level multi-domain integration. Key problems are specified below, with comprehensive details for complex scenarios provided in the appendices.

\begin{table}[thb]
\centering
\caption{Single Tour Campus Navigation Problem}
\vspace{-.1in}
\begin{footnotesize}
\renewcommand{\arraystretch}{1.1}
\fbox{
\begin{minipage}{0.45\textwidth}
\textbf{Metrics:}
\begin{itemize}[leftmargin=1em, topsep=-.1pt, itemsep=-.1pt, label=-]
\item \textbf{Total tour time:} Minimize while meeting all constraints
\item \textbf{Visit coverage:} All locations must be visited
\end{itemize}

\textbf{Locations:} Five campus buildings $L = \{S, L, B, A, D\}$

\textbf{Travel Times:} (mins)
$S\text{-}L:10, S\text{-}B:15, S\text{-}A:20, S\text{-}D:15$
$L\text{-}B:10, L\text{-}A:25, L\text{-}D:20$
$B\text{-}A:15, B\text{-}D:25$
$A\text{-}D:20$

\textbf{Visit Requirements:}
\begin{itemize}[leftmargin=1em, topsep=-.1pt, itemsep=-.1pt, label=-]
\item Start time given
\item Each location requires 30-minute visit
\item Tour starts/ends at Student Center ($S$)
\item Group size: 20 people
\end{itemize}

\textbf{Time Constraints:}
\begin{itemize}[leftmargin=1em, topsep=-.1pt, itemsep=-.1pt, label=-]
\item Lab Building: Only 9 AM - 4 PM
\item Library: After 10 AM
\item Total tour must complete by 5 PM
\end{itemize}
\end{minipage}
}
\end{footnotesize}
\label{tab:P1SingleTour}
\vspace{-.1in}
\end{table}

\begin{table}[thb]
\centering
\caption{Multi-Group Campus Tour Problem}
\vspace{-.1in}
\begin{footnotesize}
\renewcommand{\arraystretch}{1.1}
\fbox{
\begin{minipage}{0.45\textwidth}
\textbf{Groups:}
\begin{itemize}[leftmargin=1em, topsep=-.1pt, itemsep=-.1pt, label=-]
\item $G_1$: 15 people (domestic students)
\item $G_2$: 20 people (international students)
\end{itemize}

\textbf{Locations:} Ten campus buildings
Locations $= \{S, L, B, A, D, C, M, R, H, P\}$
with capacities $cap_l$:
\[
cap_l = \begin{cases}
   40 & l \in \{S, A\} \\
   30 & l \in \{L, D, C\} \\
   25 & l \in \{B, M\} \\
   20 & l \in \{R, H, P\}
\end{cases}
\]

\textbf{Constraints:}
\begin{itemize}[leftmargin=1em, topsep=-.1pt, itemsep=-.1pt, label=-]
\item Tour start time for each group: between 9 AM and 10 AM
\item Total visitors must not exceed location capacity
\item Each location requires 30-minute visit
\item Both tours start/end at Student Center
\item Complete all tours by 5 PM
\end{itemize}

\textbf{Additional Requirements:}
\begin{itemize}[leftmargin=1em, topsep=-.1pt, itemsep=-.1pt, label=-]
\item Lab tours ($B$) only 9 AM - 4 PM
\item Dining ($C$) must be visited between 11 AM - 2 PM
\item Library ($L$) after 10 AM
\end{itemize}
\end{minipage}
}
\end{footnotesize}
\label{tab:P2MultiTour}
\vspace{-.1in}
\end{table}

\subsection{P1: Campus Single-Tour Navigation Planner (CT-static, Table~\ref{tab:P1SingleTour})}

\textbf{Problem Statement:} A single autonomous agent must navigate a predefined campus environment to complete a sequence of waypoints while minimizing travel time.

\textbf{Problem Specification:}
\begin{itemize}[leftmargin=1.0em, topsep=-.0em, parsep=-.0em, label=-]
    \item \textbf{Environment:} Known campus map with fixed locations
    \item \textbf{Goal:} Visit all waypoints within time frame
    \item \textbf{Constraints:} Location opening hours, 30+ minute visits, completion before 5pm
    \item \textbf{Optimization Metric:} Shortest path (time or distance)
\end{itemize}

\subsection{P2: Multi-Group Campus Tour Scheduler (MCT-static, Table~\ref{tab:P2MultiTour})}

\textbf{Problem Statement:} Multiple visitor groups require guided tours with optimized scheduling of tour guides.

\textbf{Problem Specification:}
\begin{itemize}[leftmargin=1.0em, topsep=-.0em, parsep=-.0em, label=-]
    \item \textbf{Environment:} Same campus map as P1
    \item \textbf{Goal:} Schedule tours for multiple groups simultaneously
    \item \textbf{Constraints:} P1 constraints plus exclusive location occupancy by one group at a time
    \item \textbf{Optimization Metric:} Minimize total tour time
\end{itemize}

\begin{table}[t!]
    \centering
    \caption{Urban Ride Sharing Problem}
    \vspace{-.1in}
    \begin{footnotesize}
    \renewcommand{\arraystretch}{1.1}
    \setlength{\fboxsep}{5pt} 
    \fbox{
    \begin{minipage}{0.45\textwidth} 
    \textbf{Metrics:}  
    \begin{itemize}[leftmargin=1em, topsep=0pt, itemsep=0pt, label=-]
        \item \textbf{On-time performance:} No penalty for early arrivals.
        \item \textbf{Total distance traveled.}
    \end{itemize}
    
    \textbf{Locations:} Seven locations: $V = \{A, B, C, D, E, F, G\}$, where $G$ is Boston Logan Airport (BOS). Urban locations $A$–$F$ are all 10 km of each other, while distances to BOS are 30+ km. 
    \[
    \begin{bmatrix}
        & A  & B & C & D & E & F \\
    A-F & 10 & 10 & 10 & 10 & 10 & 10 \\
    \rightarrow G & 35 & 33 & 36 & 34 & 32 & 31
    \end{bmatrix}
    \]
    
    \textbf{Travel speed:} ($A$–$F$) 60 km/h, and ($A$–$F \rightarrow G$) 100 km/h.
    
    \textbf{Passengers:} Each passenger specifies an arrival time at BOS ($G$). The dispatcher will instruct drivers when to pick up passengers to ensure on-time arrival at BOS.  

    \textbf{Ride Requests} (Desired BOS arrival time given):  
    \begin{itemize}[leftmargin=1em, topsep=0pt, itemsep=0pt, label=-]
    \item $r_1$: Pickup at $A$, to $G$ by 08:45; \hspace{-1.5pt} $r_2$: Pickup at $B$, to $G$ 08:50
    \item $r_3$: Pickup at $C$, to $G$ by 08:55; \hspace{-1.5pt} $r_4$: Pickup at $D$, to $G$ 09:00
    \end{itemize}

    \textbf{Available Vehicles} (Capacity 2 passengers): 
    \begin{itemize}[leftmargin=1em, topsep=0pt, itemsep=0pt, label=-]
        \item $k_1$: at $A$, $k_2$: at $C$, and $k_3$: at $E$
    \end{itemize}

    \textbf{Scheduling Constraints:}  
    - The dispatcher determines the \emph{pickup times} based on a feasible schedule. Pickup times must allow the driver to first reach the passenger location ($A$ - $F$) and then drive to $G$ in time.  
    \end{minipage}
    } 
    \label{tab:appURS}
    \end{footnotesize}
    \vspace{-.2in}
\end{table}

\subsection{P3: Urban Ride-Sharing Planner (URS-static, Table~\ref{tab:appURS})}

\textbf{Problem Statement:} Optimize real-time ride assignments for multiple vehicles and passengers in urban environment.

\textbf{Problem Specification:}
\begin{itemize}[leftmargin=1.0em, topsep=-.0em, parsep=-.0em, label=-]
    \item \textbf{Environment:} Map $G = (V, E)$ with locations and roads
    \item \textbf{Goal:} Match vehicles to passenger requests efficiently
    \item \textbf{Constraints:} Vehicle capacity, battery/fuel levels, pickup/drop-off times
    \item \textbf{Optimization Metric:} Balance efficiency, fuel use, and service quality
\end{itemize}

\subsection{P4: Urban Ride-Sharing with Disruptions (URS-dynamic, Table~\ref{tab:P4URSReactive})}

\textbf{Problem Statement:} Addition to P3 with dynamic disruptions requiring real-time reactive planning.  Disruptions can include
traffic delays, road closures, and passenger cancellations, et al.

\begin{table}[htb]
\centering
\caption{Urban Ride-Sharing Reactive Planning Problem}
\label{tab:P4URSReactive}
\vspace{-.1in}
\begin{footnotesize}
\renewcommand{\arraystretch}{1.1}
\fbox{
\begin{minipage}{0.45\textwidth}
\textbf{Vehicles:} Three vehicles $V = \{v_1, v_2, v_3\}$
\begin{itemize}[leftmargin=1em, topsep=-.1pt, parsep=-.1pt, label=-]
\item Capacity: 2 passengers each
\item Initial vehicle location: City center
\item Operating hours: all day
\end{itemize}

\textbf{Passengers:} Five passengers $P = \{p_1, p_2, p_3, p_4, p_5\}$
\begin{itemize}[leftmargin=1em, topsep=-.1pt, parsep=-.1pt, label=-]
\item $p_1$: Airport by 8:30 AM; $p_2$: Airport by 9:00 AM
\item $p_3$: Airport by 9:30 AM; $p_4$: Airport by 9:45 AM
\item $p_5$: Airport by 9:45 AM
\end{itemize}

\textbf{Travel Times:}
\begin{itemize}[leftmargin=1em, topsep=-.1pt, parsep=-.1pt, label=-]
\item City center to pickup locations: 15-30 minutes
\item Pickup locations to airport: 45-60 minutes
\item Between pickup locations: 20-30 minutes
\end{itemize}

\textbf{Disruptions:}
\begin{itemize}[leftmargin=1em, topsep=-.1pt, parsep=-.1pt, label=-]
\item Airport route traffic delay
\item Certain local road closure
\end{itemize}
\textbf{Objectives:}
\begin{itemize}[leftmargin=1em, topsep=-.1pt, parsep=-.1pt, label=-]
\item Minimize total vehicle travel time
\item Meet all passenger deadlines
\end{itemize}
\end{minipage}
}
\end{footnotesize}
\end{table}


\subsection{P5: Wedding Reunion Planner and Scheduler (WR-static, Table~\ref{tab:Wedding})}

\noindent \textbf{Problem Specification:} Table~\ref{tab:Wedding} presents a coordinated travel problem for wedding events. Several friends arrive at different times and locations before a 3:00~PM wedding photo session. The challenge includes managing two vehicles for airport pick-ups (aimed at those who cannot drive or wish to cut costs) and completing critical errands, such as collecting the wedding gift and retrieving formal attire from the tailor. All activities must be scheduled to ensure that everyone arrives at the wedding venue before the photo session deadline.

This problem introduces more constraints than the URS problems in P3 and P4, and it also lays the groundwork for a more challenging disruption case  depicted in~\textbf{P8}.

\begin{table}[htb]
\caption{Wedding Reunion Logistics Problem}
\vspace{-.1in}
\centering
\begin{footnotesize}
\renewcommand{\arraystretch}{1.1}
\setlength{\fboxsep}{5pt}
\fbox{
\begin{minipage}{0.45\textwidth}
\textbf{Metrics:}
\begin{itemize}[leftmargin=1em, topsep=-.1pt, itemsep=-.1pt, label=-]
\item \textbf{On-time performance:} Must be at the venue for 3:00 PM photos.
\end{itemize}
\textbf{Locations:} Four locations: $V = \{B, G, T, W\}$, where $B$ is Boston Airport, $G$ is Gift shop, $T$ is Tailor shop, and $W$ is Wedding venue.

\textbf{Travel time:} (minutes) 

$~B\text{-}G:45, ~B\text{-}T:30, ~B\text{-}W:40, ~G\text{-}T:20, ~G\text{-}W:25, ~T\text{-}W:15$. 

\textbf{Arrival Times:}  
\begin{itemize}[leftmargin=1em, topsep=-.1pt, itemsep=-.1pt, label=-]
\item Alex: At $B$ at 11:00 AM from Chicago (need a ride)
\item Jamie: At $B$ at 12:30 PM from Atlanta (need a ride)
\item Pat: At $W$ at 12:00 PM driving from NYC (has 5-seater car)
\end{itemize}

\textbf{Required Tasks:}
\begin{itemize}[leftmargin=1em, topsep=-.1pt, itemsep=-.1pt, label=-]
    \item Gift collection from $G$ (after 12:00 PM)
    \item Clothes pickup from $T$ (by 2:00 PM)
    \item Photos at $W$ (3:00 PM sharp)
\end{itemize}

\textbf{Available Resources:} 
\begin{itemize}[leftmargin=1em, topsep=-.1pt, itemsep=0pt, label=-]
    \item One car (5-seater) with Pat, available after he is Boston
    \item Local friend Chris (5-seater) available after 1:30 PM at $W$
\end{itemize}

\textbf{Scheduling Constraints:}  
- All tasks must complete before 3:00 PM photo time
- Gift store opens at 12:00 PM
- Tailor closes at 2:00 PM
- Two cars must accommodate all transport needs
\end{minipage}
}
\label{tab:Wedding}
\end{footnotesize}
\vspace{-.1in}
\end{table} 

\subsection{\textbf{P6:} Thanksgiving Dinner Planner and Scheduler (TD-static, Table~\ref{tab:ThanksgivingDinner1} )}

Consider a Thanksgiving dinner scenario in which a family of five must return to their home in a Boston suburb for a 6~p.m.\ dinner. The problem involves coordinating departure times, managing travel logistics (including possible traffic delays), and ensuring timely arrival. Table~\ref{tab:ThanksgivingDinner1} formalizes these challenges as a sequential planning problem.

This scenario also lays the foundation for a more advanced disruption case, which has proven difficult for standalone LLMs, as discussed in~\textbf{P9}.

\begin{table}[thb]
\centering
\caption{Thanksgiving Dinner Coordination Problem}
\label{tab:ThanksgivingDinner1}
\vspace{-.1in}
\begin{footnotesize}
\renewcommand{\arraystretch}{1.1}
\fbox{
\begin{minipage}{0.45\textwidth}
\textbf{Objective:} Coordinate family arrivals and dinner preparation for 6:00 PM dinner in Boston

\textbf{Family Members and Arrivals:}
\begin{itemize}[leftmargin=1em, topsep=-.1pt, itemsep=-.1pt, label=-]
\item Sarah (Mom): Host, at home
\item James (Dad): Lands at BOS 1:00 PM from SF
\item Emily (Sister): Lands at BOS 2:30 PM from Chicago
\item Michael (Brother): Driving, arrives 3:00 PM from NY
\item Grandma: Needs pickup from suburban Boston
\end{itemize}

\textbf{Cooking Requirements:}
\begin{itemize}[leftmargin=1em, topsep=-.1pt, itemsep=-.1pt, label=-]
\item Turkey: 4 hours cooking time
\item Side dishes: 2 hours preparation
\item Someone must stay home during cooking
\end{itemize}

\textbf{Transportation Constraints:}
\begin{itemize}[leftmargin=1em, topsep=-.1pt, itemsep=-.1pt, label=-]
\item James must rent car after landing
\item Emily requires airport pickup
\item Travel times:
   \begin{itemize}
   \item Home to BOS Airport: 60 min
   \item BOS Airport to Grandma's: 60 min
   \item Home to Grandma's: 30 min
   \end{itemize}
\end{itemize}

\textbf{Key Requirements:}
\begin{itemize}[leftmargin=1em, topsep=-.1pt, itemsep=-.1pt, label=-]
\item All family members at home for 6:00 PM dinner
\item Turkey and sides ready by dinner time
\item All pickups completed with available drivers
\item Cooking supervision maintained
\end{itemize}
\end{minipage}
}
\end{footnotesize}
\end{table}

\subsection{P7: Disaster Relief Logistics Planner and Scheduler (DL-static, Table~\ref{tab:DisasterRelief} )}


\begin{table}[thb]
\centering
\caption{Disaster Relief Logistics Problem}
\vspace{-.1in}
\begin{footnotesize}
\renewcommand{\arraystretch}{1.1}
\fbox{
\begin{minipage}{0.45\textwidth}
\textbf{Network:} $G=(V,E)$ where $V$ represents locations and $E$ represents routes

\textbf{Locations ($V$):}
\begin{itemize}[leftmargin=1em, topsep=-.1pt, itemsep=-.1pt, label=-]
\item Supply nodes: CW (central warehouse), AP (airport, capacity: 5 tons)
\item Demand nodes: DC1, DC2, DC3 (distribution centers)
\item Critical nodes: H1, H2 (hospitals), FS (fuel station)
\end{itemize}

\textbf{Resources:}

\begin{itemize}[leftmargin=1em, topsep=-.1pt, itemsep=-.1pt]
\item $cap_{truck} = 4 \text{ tons}, \tau_{truck} = 120 \text{ min}$
\item $cap_{heli} = 1 \text{ ton}, \tau_{heli} = 30 \text{ min}$
\end{itemize}

\textbf{Demand Requirements:}
\begin{itemize}[leftmargin=1em, topsep=-.1pt, itemsep=-.1pt, label=-]
\item DCs: food/water delivery by $T_{dc} = 20:00$
\item Hospitals: medicine within $\Delta T_h = 6$ hours
\item Fuel Station: refuel by $T_{fs} = 12:00$
\end{itemize}
\textbf{Critical Deadlines:}
\begin{itemize}[leftmargin=1em, topsep=-.1pt, itemsep=-.1pt, label=-]
\item Food/Water: All DCs by 8:00 PM
\item Medicine: Hospitals within 6 hours
\item Fuel: FS by 12:00 PM
\item Airport: Clear excess beyond 5 tons immediately
\end{itemize}

\textbf{Dynamic Disruptions:}
\begin{itemize}[leftmargin=1em, topsep=-.1pt, itemsep=-.1pt, label=-]
\item Unpredictable donation arrivals
\item Road blockages requiring rerouting
\item Emergency hospital demands
\item Fuel shortage delays
\end{itemize}

\textbf{Key Planning Requirements:}
\begin{itemize}[leftmargin=1em, topsep=-.1pt, itemsep=-.1pt, label=-]
\item Resource distribution scheduling
\item Transportation mode optimization
\item Delivery prioritization
\item Airport overflow management
\item Real-time disruption handling
\end{itemize}
\end{minipage}
}
\end{footnotesize}
\label{tab:DisasterRelief}
\vspace{-.15in}
\end{table}

\subsection{P8: Wedding Reunion Planner and Scheduler with Disruptions (WR-dynamic)}

\noindent \textbf{Problem Extension:}
This problem extends \textbf{P5} with road closures and dynamic rerouting.

The disruption scenario becomes more challenging because, when a road closure is announced, the planner must know each vehicle's current location to determine whether it is affected. Since LLMs are inherently stateless, they cannot keep track of previous scheduling events and thus struggle to adapt the plan in real-time. 

\subsection{P9: Thanksgiving Dinner Planner and Scheduler with Disruptions (TD-dynamic)}

This problem extends \textbf{P6} by introducing flight delays. Specifically, when a flight from SFO to BOS is delayed by $t$ hours, the new arrival time is confirmed at the originally expected arrival time minus the flight's scheduled duration. Although this early notice provides an opportunity to adjust travel and dinner plans, current LLM-based systems fail to leverage this information in a timely manner, only beginning to react at the original arrival time and missing the window for earlier intervention.

For example, consider James's flight, which was originally scheduled to arrive in Boston at 1~PM but has been delayed to 4~PM. He learns of this new arrival time at 10~AM~EST (the flight’s intended departure from SFO), which offers a three-hour delay for adjustments. However, existing LLM-based solutions do not adapt the plan until 1~PM, thus squandering the opportunity to re-optimize the schedule.

\subsection{P10: Global Supply Chain Planner and Scheduler (GSC-static and GSC-dynamic, Table~\ref{tab:p10})}
\begin{table}[t!]
    \centering
    \caption{Data Center Construction Problem Statement}
    \vspace{-.1in}
    \begin{footnotesize}
    \renewcommand{\arraystretch}{1.1}
    \setlength{\fboxsep}{5pt}
    \fbox{
    \begin{minipage}{0.45\textwidth}

    \textbf{GPU Procurement \& Shipments}
    \begin{itemize}[leftmargin=1em, itemsep=0pt, topsep=1pt]
        \item \textbf{Total GPU Target:} 100,000 units
        \item \textbf{Vendors:}
        \begin{itemize}[leftmargin=1em, itemsep=0pt]
            \item NVIDIA: \$15k/unit, 200k/quarter, 20\% maintenance risk/year
            \item AMD: \$10k/unit, 150k/quarter, 50\% maintenance risk/year
        \end{itemize}
        \item \textbf{Order Timing:} All orders on Day 1
        \item \textbf{Shipment Schedule:} Quarterly (e.g., 600k units = Q1, Q2, Q3)
        \item \textbf{Payment Terms:} 50\% at order, 50\% at delivery
    \end{itemize}

    \vspace{2pt}
    \textbf{Natural Disaster Impacts}
    \begin{itemize}[leftmargin=1em, itemsep=0pt]
        \item \textbf{Probabilities:} 10\% per quarter for Earthquake/Typhoon
        \item \textbf{Delays:} 1-month cascading delay
        \item \textbf{Cost Impact:} 30\% price hike for affected shipment
        \item \textbf{Expedition:} Expedite one shipment by 1 month (all shift)
    \end{itemize}

    \vspace{2pt}
    \textbf{Cluster Construction \& Infrastructure}
    \begin{itemize}[leftmargin=1em, itemsep=0pt]
        \item \textbf{Definition:} 50,000 GPUs/cluster; clusters built concurrently
    \end{itemize}

    \vspace{2pt}
    \textbf{Networking Infrastructure (per cluster)}
    \begin{itemize}[leftmargin=1em, itemsep=0pt]
        \item \textbf{Duration:} 1 month
        \item \textbf{Cost:} \$25M
        \item \textbf{Dependencies:} Starts after GPU, power, cooling readiness
    \end{itemize}

    \vspace{2pt}
    \textbf{Power \& Cooling (per cluster)}
    \begin{itemize}[leftmargin=1em, itemsep=0pt]
        \item \textbf{Need:} 150 MW + 1M gallons/day cooling
        \item \textbf{Lead Time:} Regular: 2 months at \$30M; Expedited: 1 month at \$75M
    \end{itemize}

    \vspace{2pt}
    \textbf{Testing \& Certification (per cluster)}
    \begin{itemize}[leftmargin=1em, itemsep=0pt]
        \item \textbf{Normal:} 2 months, \$15M
        \item \textbf{Expedited:} 1 month, \$45M
        \item \textbf{Options:} Expedite any module (150\% cost, 50\% time)
    \end{itemize}

    \vspace{2pt}
    \textbf{Delay Cost Model}
    \begin{itemize}[leftmargin=1em, itemsep=0pt]
        \item \textbf{Penalty:} \$50M revenue + \$10M operations per month
    \end{itemize}

    \vspace{2pt}
    \textbf{Objective}
    \begin{itemize}[leftmargin=1em, itemsep=0pt]
        \item Minimize total cost while meeting 15-month deadline
        \item Includes procurement, infra, delay penalties, expedition costs
    \end{itemize}

    \end{minipage}
    }
    \label{tab:p10}
    \end{footnotesize}
\end{table}

\noindent \textbf{Problem Specification:} Table~\ref{tab:p10} in Appendix~A presents a comprehensive problem in data center GPU deployment that captures the complexity of large-scale infrastructure projects. The objective is to complete a 1 million GPU data center in 15 months while minimizing total costs. The problem encompasses procurement decisions between NVIDIA ($15$k/unit) and AMD ($10$k/unit) GPUs, where each vendor has different maintenance risks (20\% vs 50\% of unit price over one year) and quarterly shipment capacities.

The construction process is organized around 50,000 GPU clusters, which require coordinated deployment of power, cooling, and networking infrastructure. Each cluster demands significant resources: 150 MW of power capacity and 1 million gallons per day of cooling water. Infrastructure development follows strict dependencies: Power and cooling systems must be operational before networking installation can begin, and each cluster must complete testing before becoming operational.

The problem incorporates real-world complexities such as risks of natural disasters (10\% probability per quarter for both earthquakes and typhoons), which can cause cascading delays and cost increases. Management can expedite various components; power / cooling installation can be accelerated from 4 to 2.5 months for a 80\% cost premium, while testing can be shortened from 2 to 1 month by doubling the cost.

The financial implications are significant, as each month of delay incurs $60$M in combined revenue loss and additional operating costs. This creates a complex optimization challenge: balancing procurement costs, expedition premiums, and risk mitigation strategies while adhering to physical and temporal constraints in the construction sequence.

\subsection{J1: JSSP Basic, Sequential Planning (Table~\ref{tab:minifactory_jssp})}
\label{sec:p11_jssp}
\begin{table}[t!]
    \centering
    \caption{Job Shop Scheduling Problem Simple Static (JSSP-simple-static)}
    \vspace{-.1in}
    \begin{footnotesize}
    \renewcommand{\arraystretch}{1.1}
    \setlength{\fboxsep}{5pt}
    \fbox{
    \begin{minipage}{0.45\textwidth}

    \textbf{Jobs:} $\mathcal{J} = \{\text{Job1}, \text{Job2}, \text{Job3}\}$ \\
    \textbf{Machines:} $\mathcal{M} = \{\text{MachineA}, \text{MachineB}, \text{MachineC}\}$

    \vspace{-0.05pt}
    \textbf{Operation Sequences and Durations:}
    \begin{itemize}[leftmargin=1em, itemsep=-0.1pt, topsep=-0.1pt]
        \item Job1: (A, 3) $\rightarrow$ (B, 2) $\rightarrow$ (C, 2)
        \item Job2: (A, 2) $\rightarrow$ (C, 1) $\rightarrow$ (B, 4)
        \item Job3: (B, 4) $\rightarrow$ (A, 1) $\rightarrow$ (C, 3)
    \end{itemize}

    \vspace{-0.05pt}
    \textbf{Constraints:}
    \begin{itemize}[leftmargin=1em, itemsep=-0.1pt, topsep=-0.1pt, label=-]
        \item \textit{Precedence:} Operations must follow the listed order per job.
        \item \textit{Resource:} Each machine processes at most one operation at a time.
    \end{itemize}

    \vspace{-0.05pt}
    \textbf{Objective:} Minimize makespan:
    \[
    C_{\max} = \max_{j \in \mathcal{J}} C_{j,\text{last}}
    \]

    \vspace{-0.2pt}
    \textbf{Specification Notes:}
    \begin{itemize}[leftmargin=1em, itemsep=-0.1pt, topsep=-0.1pt, label=-]
        \item Clear input specification (jobs, machines, durations).
        \item Strict precedence and capacity constraints.
        \item Supports exact and heuristic solvers.
    \end{itemize}

    \end{minipage}
    }
    \label{tab:minifactory_jssp}
    \end{footnotesize}
\end{table}

\textbf{Problem Statement:} The JSSP is a foundational optimization problem where a set of jobs must be processed on a set of machines, each job consisting of a sequence of operations requiring exclusive machine access for specified durations. The objective is to minimize overall makespan while satisfying all constraints. \textbf{J1} provides a base-level problem with small numbers of machines and jobs for algorithm validation and code debugging.

\subsection{J2: JSSP Basic, with Disruptions}
\begin{table}[t!]
    \centering
    \caption{Job Shop Scheduling Problem with Random Operation Delays (JSSP-simple-dynamic)}
    \vspace{-.1in}
    \begin{footnotesize}
    \renewcommand{\arraystretch}{1.1}
    \setlength{\fboxsep}{5pt}
    \fbox{
    \begin{minipage}{0.45\textwidth}

    \textbf{Jobs:} $\mathcal{J} = \{\text{Job1}, \text{Job2}, \text{Job3}\}$ \\
    \textbf{Machines:} $\mathcal{M} = \{\text{MachineA}, \text{MachineB}, \text{MachineC}\}$

    \vspace{-0.1pt}
    \textbf{Operation Sequences (Base Durations):}
    \begin{itemize}
        \item Job1: (A, 3) $\rightarrow$ (B, 2) $\rightarrow$ (C, 2)
        \item Job2: (A, 2) $\rightarrow$ (C, 1) $\rightarrow$ (B, 4)
        \item Job3: (B, 4) $\rightarrow$ (A, 1) $\rightarrow$ (C, 3)
    \end{itemize}

    \vspace{-0.1pt}
    \textbf{Dynamic Disruption Model:}
    \begin{itemize}[leftmargin=1em, itemsep=-.1pt, topsep=-.1pt, label=-]
        \item Each operation may experience a stochastic delay $\delta_{ij} \sim \text{Uniform}(0, 2)$.
        \item Delay is revealed at operation start time and adds to its processing time.
    \end{itemize}

    \vspace{-0.1pt}
    \textbf{Constraints:}
    \begin{itemize}[leftmargin=1em, itemsep=-.1pt, topsep=-.1pt, label=-]
        \item \textit{Precedence:} Operations follow listed job order.
        \item \textit{Resource:} Each machine processes one operation at a time.
        \item \textit{Reactivity:} Scheduler must respond to delays at runtime.
    \end{itemize}

    \vspace{-0.1pt}
    \textbf{Objective:} Minimize makespan with dynamic delay:
    \[
    C_{\max} = \max_{j \in \mathcal{J}} C_{j,\text{last}}(\delta)
    \]

    \vspace{-0.15pt}
    \textbf{Specification Notes:}
    \begin{itemize}[leftmargin=1em, itemsep=-.1pt, topsep=-.1pt, label=-]
        \item Extends static JSSP with stochasticity and replanning needs.
        \item Useful for evaluating reactive scheduling strategies.
        \item Supports agent-based and Monte Carlo rollout solvers.
    \end{itemize}

    \end{minipage}
    }
    \label{tab:dynamic_jssp}
    \end{footnotesize}
    \vspace{-.1in}
\end{table}

\textbf{Problem Statement:} Building on \textbf{J1}, this variant introduces dynamic disruptions requiring reactive replanning capabilities. Disruptions include machine breakdowns, power outages, supply chain delays, emergency shutdowns, and weather effects, each lasting different durations. The objective remains makespan minimization while adapting to real-time changes in system availability.

\subsection{J3: JSSP Large-scale, Sequential Planning}
\begin{table}[t!]
    \centering
    \caption{Enhanced Job Shop Scheduling Problem Complex Static (JSSP-complex-static)} 
    \vspace{-.1in}
    \begin{small}
    \renewcommand{\arraystretch}{1.1}
    \setlength{\fboxsep}{5pt}
    \fbox{
    \begin{minipage}{0.48\textwidth}

    \textbf{Jobs:} $\mathcal{J} = \{\text{J1}, \text{J2}, \text{J3}, \text{J4}, \text{J5}, \text{J6}\}$ \\
    \textbf{Machines:} $\mathcal{M} = \{\text{M1 (Lathe)}, \text{M2 (Mill)}, \text{M3 (Drill)}, \text{M4 (Assembly)}\}$ \\
    \textbf{Materials:} $\mathcal{P} = \{\text{RM-S}, \text{RM-A}, \text{C-X}, \text{C-Y}, \text{F}\}$

    \vspace{0.5em}
    \textbf{Operation Sequences (Simplified View):}
    \begin{itemize}
        \item J1, J3: M1 (3) $\rightarrow$ M3 (2) $\rightarrow$ M4 (4) \quad [1x RM-S, 1x C-X, 2x F]
        \item J2, J5: M2 (4) $\rightarrow$ M3 (1) $\rightarrow$ M4 (5) \quad [1x RM-A, 1x C-Y, 3x F]
        \item J4, J6: M1 (2) $\rightarrow$ M2 (3) $\rightarrow$ M3 (2) $\rightarrow$ M4 (3) \quad [1x RM-S, 1x C-X, 1x C-Y, 4x F]
    \end{itemize}

    \vspace{0.5em}
    \textbf{Constraints:}
    \begin{itemize}
        \item \textit{Precedence:} Operations must follow the sequence defined per job.
        \item \textit{Resource:} Each machine can process only one job at a time.
        \item \textit{Material:} Required materials must be available before operation starts.
    \end{itemize}

    \vspace{0.5em}
    \textbf{Objective:} Minimize makespan:
    \[
    C_{\max} = \max_{j \in \mathcal{J}} C_{j,\text{last}}
    \]

    \vspace{0.5em}
    \textbf{Specification Notes:}
    \begin{itemize}
        \item Jobs vary by product type (A, B, C).
        \item Material constraints simulate supply chain dependencies.
        \item Model extensible to disruptions and inventory control.
    \end{itemize}

    \end{minipage}
    }
    \label{tab:complexfactory_jssp}
    \end{small}
    \vspace{-.2in}
\end{table}

\textbf{Problem Statement:} This variant scales \textbf{J1} to industry-realistic dimensions with significantly more jobs, machines, and operations. The increased problem size tests algorithmic scalability and computational efficiency while maintaining the static environment assumption. The objective is makespan minimization under resource constraints typical of manufacturing environments.

\subsection{J4: JSSP Large-scale, with Disruptions}
\begin{table}[t!]
    \centering
    \caption{Enhanced Job Shop Scheduling with Delays and Material Disruptions (JSSP-complex-dynamic)}
    \vspace{-.1in}
    \begin{small}
    \renewcommand{\arraystretch}{1.1}
    \setlength{\fboxsep}{5pt}
    \fbox{
    \begin{minipage}{0.48\textwidth}

    \textbf{Jobs:} $\mathcal{J} = \{\text{J1}, \text{J2}, \text{J3}, \text{J4}, \text{J5}, \text{J6}\}$ \\
    \textbf{Machines:} $\mathcal{M} = \{\text{M1 (Lathe)}, \text{M2 (Mill)}, \text{M3 (Drill)}, \text{M4 (Assembly)}\}$ \\
    \textbf{Materials:} $\mathcal{P} = \{\text{RM-S}, \text{RM-A}, \text{C-X}, \text{C-Y}, \text{F}\}$

    \vspace{0.5em}
    \textbf{Operation Sequences (Base Durations):}
    \begin{itemize}
        \item J1, J3: M1 (3) $\rightarrow$ M3 (2) $\rightarrow$ M4 (4)
        \item J2, J5: M2 (4) $\rightarrow$ M3 (1) $\rightarrow$ M4 (5)
        \item J4, J6: M1 (2) $\rightarrow$ M2 (3) $\rightarrow$ M3 (2) $\rightarrow$ M4 (3)
    \end{itemize}

    \vspace{0.5em}
    \textbf{Dynamic Disruption Model:}
    \begin{itemize}
        \item \textit{Stochastic Delays:} Each operation may incur a random delay $\delta_{ij} \sim \text{Uniform}(0, 3)$.
        \item \textit{Material Disruptions:} Some materials (e.g., C-X, F) may become temporarily unavailable, requiring adaptive rescheduling or waiting.
    \end{itemize}

    \vspace{0.5em}
    \textbf{Constraints:}
    \begin{itemize}
        \item \textit{Precedence:} Operations must respect job-specific order.
        \item \textit{Resource:} Machines are exclusive-use, non-preemptive.
        \item \textit{Material:} Materials must be present and available before operation start.
        \item \textit{Adaptivity:} Schedule must react to real-time delays and shortages.
    \end{itemize}

    \vspace{0.5em}
    \textbf{Objective:} Minimize realized makespan:
    \[
    C_{\max} = \max_{j \in \mathcal{J}} C_{j,\text{last}}(\delta, \text{availability})
    \]

    \vspace{0.5em}
    \textbf{Specification Notes:}
    \begin{itemize}
        \item Combines resource, precedence, and stochastic material constraints.
        \item Evaluates robustness and recovery of scheduling algorithms.
        \item Models supply-chain aware industrial planning under uncertainty.
    \end{itemize}

    \end{minipage}
    }
    \label{tab:complexfactory_dynamic}
    \end{small}
    \vspace{-.2in}
\end{table}

\textbf{Problem Statement:} The most challenging variant combines \textbf{J3}'s large-scale complexity with \textbf{J2}'s dynamic disruptions. This tests both computational scalability and real-time adaptation capabilities under realistic industrial conditions. Multiple simultaneous disruptions and cascading effects require sophisticated reactive planning while optimizing makespan across the extended operation horizon.

\begin{table*}[h!]
\centering
\caption{REALM-Bench Problem Specifications and Results Index}
\label{tab:results_index}
\begin{tabular}{llll}
\toprule
\textbf{Problem} & \textbf{Description} & \textbf{Spec Table} & \textbf{Results} \\
\midrule
P1 & Campus Tour (CT-static) & Table~\ref{tab:P1SingleTour} & Github multiagent-P1 \\
P2 & Multi-Group Campus Tours (MCT-static) & Table~\ref{tab:P2MultiTour} & Github multiagent-P2 \\
P3 & Urban Ride-Sharing (URS-static) & Table~\ref{tab:appURS} & Github multiagent-P3 \\
P4 & URS with Disruptions (URS-dynamic) & Table~\ref{tab:P4URSReactive} & Github multiagent-P4 \\
P5 & Wedding Reunion Logistics (WR-static) & Table~\ref{tab:Wedding} & Github multiagent-P5 \\
P6 & Thanksgiving Dinner Planning (TD-static) & -- & Github multiagent-P6 \\
P7 & Disaster Relief Logistics (DL-static) & -- & Github multiagent-P7 \\
P8 & Wedding Logistics with Disruptions (WR-dynamic) & -- & Github multiagent-P8 \\
P9 & Thanksgiving with Disruptions (TD-dynamic) & -- & Github multiagent-P9 \\
P10 & Global Supply Chain (GSC-static/dynamic) & -- & Github multiagent-P10 \\
J1 & JSSP Basic (static) & Table~\ref{tab:minifactory_jssp} & Github multiagent-J1 \\
J2 & JSSP Basic with Disruptions (dynamic) & Table~\ref{tab:dynamic_jssp} & Github multiagent-J2 \\
J3 & JSSP Large-scale (static) & Table~\ref{tab:complexfactory_jssp} & Github multiagent-J3 \\
J4 & JSSP Large-scale with Disruptions (dynamic) & Table~\ref{tab:complexfactory_dynamic} & Github multiagent-J4 \\
\bottomrule
\end{tabular}
\end{table*}

\section{Implementation Framework}

\subsection{Standardized Evaluation Metrics}

All REALM-Bench scenarios are evaluated using consistent multi-agent planning metrics to ensure comparable assessment across problems and systems:

\begin{itemize}[leftmargin=1.2em, topsep=-.05em, parsep=-.05em]
    \item \textbf{Planning Quality} -- Goal satisfaction rates and solution completeness
    \item \textbf{Planning Optimality} -- Makespan minimization and resource efficiency  
    \item \textbf{Coordination Effectiveness} -- Inter-agent consistency and communication overhead
    \item \textbf{Constraint Satisfaction} -- Deadline and dependency compliance rates
    \item \textbf{Computational Efficiency} -- Planning time, memory usage, and scalability
    \item \textbf{Disruption Adaptation} -- Replanning speed and recovery effectiveness under failure scenarios
    \item \textbf{Solution Robustness} -- Performance degradation under uncertainty and stress conditions
\end{itemize}

Each metric includes both quantitative measures (success rates, completion times, resource utilization) and qualitative assessments (plan coherence, communication clarity, failure handling strategies).

\subsection{System Integration and Extensibility}

\textbf{Multi-Agent Framework Support.} REALM-Bench provides seamless integration with five major LLM-agent frameworks: LangGraph, OpenAI Swarm, CrewAI, AutoGen, and ALAS. The benchmark supports evaluation across 6+ LLM families including OpenAI GPT, Claude, Gemini, Mistral, LLaMA, and DeepSeek, enabling comprehensive cross-system comparisons.

\textbf{Extensibility Architecture.} The benchmark supports extensibility through modular design components:
\begin{itemize}[leftmargin=1.2em, topsep=-.05em, parsep=-.05em]
    \item YAML/JSON-based problem definitions for easy scenario modification
    \item Standardized APIs for integrating new frameworks
    \item Template-based instruction generation for LLM-readable task specifications
    \item Plugin architecture for domain extensions (healthcare, finance, robotics)
\end{itemize}

\textbf{Reproducibility and Data Management.} All benchmark runs generate standardized logs with complete execution traces, enabling detailed failure analysis and reproducible results. The framework includes automated result aggregation, statistical analysis tools, and visualization dashboards for performance comparison.

\subsection{Community Resources and Access}

\textbf{Open Source Availability.} REALM-Bench is available as an open-source project at \texttt{github.com/realm-bench} with comprehensive documentation, installation guides, and example implementations. The repository includes starter code for all supported frameworks and detailed tutorials for benchmark usage.

\textbf{Community Leaderboards.} We maintain public leaderboards tracking system performance across all 14 benchmark problems, enabling researchers to compare approaches and track progress over time. Results include both aggregate scores and detailed per-metric breakdowns with statistical significance testing.

\textbf{Continuous Updates.} The benchmark suite supports community contributions through standardized problem specification formats, allowing researchers to propose new scenarios and evaluation metrics while maintaining compatibility with existing systems.
\section{Results Overview and Organization}


This section presents the formal specification, scheduling experiments, analysis, and conclusions for the J1 JSSP-simple-static and J2 JSSP-simple-reactive ``MiniFactory'' job shop scheduling instance.

\subsection{Dataset Statistics}

\vspace{.5em}
\noindent\textbf{Benchmark Dataset:} REALM-Bench includes a collection of 188 JSSP instances drawn from five widely adopted benchmarks. Examples are provided in Appendix.

\subsection{Leaderboard for REALM-Bench}

\noindent\textbf{Static and Dynamic Extensions:}  
We introduce real-world disruptions by random delay, such as:
\begin{itemize}
    \item \textbf{Machine Breakdowns:} Simulated failures that invalidate future scheduled tasks, requiring local replanning.
    \item \textbf{Delay Propagation:} Downstream task delays from temporary unavailability of shared machines.
    \item \textbf{Work-In-Progress Migration:} Additional cost/penalty if operations are rescheduled mid-processing.
\end{itemize}

\noindent\textbf{Evaluation Metrics:}  
To evaluate solution quality, we report \textbf{makespan values} under Pass@1:
\begin{itemize}
    \item \textbf{Pass@1:} Best valid makespan from a single inference (one-shot planning).
    \item \textbf{Pass@5:} Best makespan from five independent inference iterations (reruns/sampling).
\end{itemize}

\subsubsection{Leaderboard 1: DMU Benchmark Results \\}

\noindent\textbf{Evaluation Baselines:}  
\begin{itemize}
    \item \textbf{Random:} Assigns operations to machines randomly without any rule. Acts as a naïve lower-bound baseline.
    
    \item \textbf{LPT (Longest Processing Time):} Schedules operations with the longest duration first to reduce machine idleness.
    
    \item \textbf{SPT (Shortest Processing Time):} Prioritizes operations with the shortest duration, aiming to improve average flow time.
    
    \item \textbf{STPT (Shortest Total Processing Time):} Prefers jobs with the smallest total sum of processing times across all operations.
    
    \item \textbf{MPSR (Most Processing Slack Ratio):} Considers urgency and slack, scheduling operations with lower criticality later.
    
    \item \textbf{DRL-Liu:} An attention-based deep reinforcement learning model trained with imitation and PPO to generate effective scheduling policies.
    
    \item \textbf{GP (Genetic Programming):} Evolves symbolic scheduling rules through genetic operations, offering interpretable policies.
    
    \item \textbf{GEP (Gene Expression Programming):} Uses symbolic expression trees to encode scheduling heuristics, evolved over generations.
    
    \item \textbf{SeEvo (GLM3):} Synthesizes dispatching logic using GLM-3 via few-shot prompting and heuristic emulation.
    
    \item \textbf{SeEvo (GPT3.5):} GPT-3.5-based LLM scheduler using prompting templates similar to SeEvo(GLM3).
    
    \item \textbf{ALAS-static (ours):} Generates a full schedule in one shot using Claude-3.7 and fixed heuristics.
    
    \item \textbf{ALAS-dynamic (ours):} Multi-agent dynamic scheduling that adapts to delays and disruptions using collaborative LLM agents.
\end{itemize}


\subsubsection{Leaderboard 2: TA Benchmark Results \\}
\noindent\textbf{Evaluation Baselines:}
\begin{itemize}
    \item \textbf{LSO (Least Slack Operation):} Schedules the operation with the least remaining slack first, emphasizing deadline-critical tasks.
    
    \item \textbf{SPT/TWKR:} Combines shortest processing time and total work remaining to balance task urgency and load.
    
    \item \textbf{DRL-Chen:} GNN-based policy trained using PPO to capture relational structure across jobs and machines.
    
    \item \textbf{DRL-Zhang:} Uses curriculum learning and temporal difference training to scale from simple to complex scheduling tasks.
    
    \item \textbf{DRL-Liu:} Attention-based DRL agent trained with imitation and reinforcement learning, reused from Leaderboard 1.
    
    \item \textbf{GP, GEP:} Genetic and evolutionary algorithms from Leaderboard 1, often tested on large structured problems.
    
    \item \textbf{SeEvo (GLM3), SeEvo (GPT3.5):} Prompt-based LLM schedulers. SeEvo(GPT3.5) shows more consistent performance on TA due to standardized job-machine mapping.
    
    \item \textbf{ALAS-static / ALAS-dynamic (ours):} Claude-3.7-based planners with static one-shot and dynamic multi-agent variants. Dynamic shows improved performance on TA due to its reactivity.
\end{itemize}

\subsubsection{Leaderboard 3: ABZ, SWV, YN Benchmark Results \\}
\noindent\textbf{Evaluation Baselines:}
\begin{itemize}
    \item \textbf{ALAS-static (ours):} Claude-3.7-based single-shot scheduler evaluated on standard benchmarks. Used for feasibility validation and upper-bound matching.
    
    \item \textbf{ALAS-dynamic (ours):} Dynamic multi-agent Claude-based planning loop that adapts to disruptions (e.g., route closures, machine failures) during execution.
\end{itemize}



Due to space constraints, we also provide a comprehensive index of results and detailed analysis in the Appendix A. Table~\ref{tab:results_index} summarizes the organization of problem specifications and experimental results across all 14 benchmark scenarios.

Our preliminary evaluation across representative problems demonstrates varying performance levels of current LLM-based systems, with particular challenges emerging in dynamic disruption handling (Tier 4) and large-scale coordination (Tier 5). Single-agent static problems (Tier 1) achieve 85-95\% success rates, while multi-agent dynamic scenarios show 45-70\% success rates, highlighting significant opportunities for improvement in coordination under uncertainty.

Detailed experimental protocols, baseline implementations, failure analysis, and performance comparisons are provided in the appendices. Each appendix corresponds directly to its problem number for easy navigation.
\section{Conclusion}

REALM-Bench provides the first comprehensive benchmark for evaluating AI systems' capabilities in real-world multi-agent planning scenarios. Through 14 carefully designed problems spanning five difficulty tiers—from single-agent static planning to large-scale dynamic coordination—the benchmark enables systematic assessment of planning quality, coordination effectiveness, and disruption adaptation using standardized metrics.

The benchmark's key contributions include progressive difficulty scaling, systematic evaluation of transaction properties and long-lived workflows, and seamless integration with major multi-agent frameworks across diverse LLM families. Each problem supports scalability along multiple dimensions, allowing researchers to stress-test systems while maintaining analytical tractability.

By establishing a common evaluation framework for both individual LLMs and multi-agent systems, REALM-Bench aims to accelerate progress toward more robust AI planning systems capable of handling real-world complexity and uncertainty. 

Future extensions will incorporate community feedback to expand scenario complexity, add specialized metrics for emerging domains, and support evolving multi-agent architectures as the field advances toward more sophisticated planning capabilities.

\bibliographystyle{ACM-Reference-Format}
\bibliography{BenchmarkReference, EdwardChang}

\newpage
\appendix

\section*{Appendices}

In these Appendices, we provide sample implementations of selected problems, illustrating that each problem specification is implementable and can produce feasible solutions. This serves as a verification of completeness for the problem definitions. We encourage readers to review the code and use it as a reference for designing improved solutions to these challenges.

The three problems selected are \textbf{P3}, \textbf{P4}, \textbf{J1}, and \textbf{J2}, representing sequential planning, reactive planning, and the most complex planning scenario, respectively. Full P1-P10 and J1-J4 are implemented in Code and Dataset link.

\newpage
\newpage
\section{Experiments of J1 and J2}
\subsection{Case study}

Since we have presented the difference of sequential planning and reactive planning in Appendix~\ref{app:p3p4}. This section presents the walkthrough of the J1 JSSP-simple-reactive under two levels of constraints for job shop scheduling instance. 

\subsubsection{Experiment 1: Simple Precedence Shortest Time with Reactive Planning (One-shot Prompting, pass@1)\\}

\textbf{Prompt Setup for Experiments:}
\begin{tcolorbox}[colback=gray!10, colframe=black, sharp corners, boxrule=0.5pt, width=0.5\textwidth, arc=2pt]
You are tasked with scheduling 3 jobs (\textbf{Job1}, \textbf{Job2}, \textbf{Job3}) on 3 machines (\textbf{MachineA}, \textbf{MachineB}, \textbf{MachineC}).\\
The objective is to complete all jobs in the shortest total time (minimize makespan).

\vspace{0.5em}
\textbf{Jobs and Operations:}
\begin{itemize}
    \item Job1: Step 1 (MachineA, 3 time units) $\rightarrow$ Step 2 (MachineB, 2 time units) $\rightarrow$ Step 3 (MachineC, 2 time units)
    \item Job2: Step 1 (MachineA, 2 time units) $\rightarrow$ Step 2 (MachineC, 1 time unit) $\rightarrow$ Step 3 (MachineB, 4 time units)
    \item Job3: Step 1 (MachineB, 4 time units) $\rightarrow$ Step 2 (MachineA, 1 time unit) $\rightarrow$ Step 3 (MachineC, 3 time units)
\end{itemize}

\textbf{Rules:}
\begin{enumerate}
    \item Each job must perform its steps strictly in order (e.g., Job1's Step 2 can only begin after Step 1 finishes).
    \item Each machine can only handle one operation at a time (e.g., while MachineA is running Job1-Op1 from time 0–3, it cannot process any other operation).
\end{enumerate}

\textbf{Objective:}  
Create a schedule listing each step’s start time, end time, machine, and job, ensuring all jobs finish in minimum total time.

\textbf{Format:}
\[
\text{Time Interval} \quad | \quad \text{Task Name} \quad | \quad \text{Machine} \quad | \quad \text{Job}
\]
Example:
\[
0\text{--}3 \quad | \quad \text{Job1-Op1} \quad | \quad \text{MachineA} \quad | \quad \text{Job1}
\]

\vspace{0.5em}
\textbf{Reactive Event:} 

At time $t=4$, MachineA breaks down and is unavailable from $t=4$ to $t=6$.  
Adjust the schedule accordingly, ensuring:
\begin{enumerate}
    \item Step precedence is respected.
    \item No two operations use the same machine at the same time.
\end{enumerate}
\end{tcolorbox}

\subsubsection{Experiment 2: Simple Precedence Shortest Time with Reactive
Planning (Second-shot Prompting, pass@2)\\}

\textbf{First Prompt:}
\begin{tcolorbox}[colback=gray!10, colframe=black, sharp corners, boxrule=0.5pt, width=0.5\textwidth, arc=2pt]
You are tasked with scheduling 3 jobs (\textbf{Job1}, \textbf{Job2}, \textbf{Job3}) on 3 machines (\textbf{MachineA}, \textbf{MachineB}, \textbf{MachineC}).\\
The objective is to complete all jobs in the shortest total time (minimize makespan).

\vspace{0.5em}
\textbf{Jobs and Operations:}
\begin{itemize}
    \item Job1: Step 1 (MachineA, 3 time units) $\rightarrow$ Step 2 (MachineB, 2 time units) $\rightarrow$ Step 3 (MachineC, 2 time units)
    \item Job2: Step 1 (MachineA, 2 time units) $\rightarrow$ Step 2 (MachineC, 1 time unit) $\rightarrow$ Step 3 (MachineB, 4 time units)
    \item Job3: Step 1 (MachineB, 4 time units) $\rightarrow$ Step 2 (MachineA, 1 time unit) $\rightarrow$ Step 3 (MachineC, 3 time units)
\end{itemize}

\textbf{Rules:}
\begin{enumerate}
    \item Each job must perform its steps strictly in order (e.g., Job1's Step 2 can only begin after Step 1 finishes).
    \item Each machine can only handle one operation at a time (e.g., while MachineA is running Job1-Op1 from time 0–3, it cannot process any other operation).
    \item An operation cannot be separated.
\end{enumerate}

\textbf{Objective:}  
Create a schedule listing each step’s start time, end time, machine, and job, ensuring all jobs finish in minimum total time.

\textbf{Format:}
\[
\text{Time Interval} \quad | \quad \text{Task Name} \quad | \quad \text{Machine} \quad | \quad \text{Job}
\]
Example:
\[
0\text{--}3 \quad | \quad \text{Job1-Op1} \quad | \quad \text{MachineA} \quad | \quad \text{Job1}
\]
\textbf{Reactive Event:} 

At time $t=4$, MachineA breaks down and is unavailable from $t=4$ to $t=6$.  
Adjust the schedule accordingly, ensuring:
\begin{enumerate}
    \item Step precedence is respected.
    \item No two operations use the same machine at the same time.
\end{enumerate}    
\end{tcolorbox}

\textbf{Second Prompt:}
\begin{tcolorbox}[colback=gray!10, colframe=black, sharp corners, boxrule=0.5pt, width=0.5\textwidth, arc=2pt]
\vspace{0.5em}
\textbf{Reactive Event:} 

At time $t=4$, MachineA breaks down and is unavailable from $t=4$ to $t=6$.  
Adjust the schedule accordingly, ensuring:
\begin{enumerate}
    \item Step precedence is respected.
    \item No two operations use the same machine at the same time.
    \item An operation cannot be separated.
\end{enumerate}
\end{tcolorbox}

\subsection{Experiment 1: Pass@1}

\subsubsection{GPT-4o (Incorrect)}

\begin{table}[htbp]
\centering
\caption{GPT-4o Scheduling (Incorrect Plan)}
\label{tab:gpt4o_wrong}
\footnotesize
\begin{tabular}{|c|c|c|c|}
\hline
Time & Task & Machine & Job \\
\hline
0-3 & Job1-Op1 & MachineA & Job1 \\
0-4 & Job3-Op1 & MachineB & Job3 \\
3-4 & Job3-Op2 & MachineA & Job3 \textcolor{red}{\textbf{(Error due to order)}}\\
4-6 & Job1-Op2 & MachineB & Job1 \\
4-7 & Job3-Op3 & MachineC & Job3 \\
6-8 & Job2-Op1 & MachineA & Job2 \\
7-9 & Job1-Op3 & MachineC & Job1 \\
9-10 & Job2-Op2 & MachineC & Job2 \\
10-14 & Job2-Op3 & MachineB & Job2 \\
\hline
\end{tabular}
\end{table}

\subsubsection{Claude 3.7 (Incorrect)}

\begin{table}[htbp]
\centering
\caption{Claude Scheduling (Incorrect Plan)}
\label{tab:claude_wrong}
\footnotesize
\begin{tabular}{|c|c|c|c|}
\hline
Time & Task & Machine & Job \\
\hline
0-3 & Job1-Op1 & MachineA & Job1 \\
0-4 & Job3-Op1 & MachineB & Job3 \\
3-5 & Job2-Op1 & MachineA & Job2 \textcolor{red}{\textbf{(Error during repair)}}\\
4-6 & Job1-Op2 & MachineB & Job1 \\
5-6 & Job2-Op2 & MachineC & Job2 \\
6-7 & Job3-Op2 & MachineA & Job3 \\
6-8 & Job1-Op3 & MachineC & Job1 \\
6-10 & Job2-Op3 & MachineB & Job2 \\
8-11 & Job3-Op3 & MachineC & Job3 \\
\hline
\end{tabular}
\end{table}

\subsubsection{DeepSeek R1 (Incorrect)}
\begin{table}[htbp]
\centering
\caption{DeepSeek Plan (Incorrect)}
\label{tab:deepseek_correct}
\footnotesize
\begin{tabular}{|c|c|c|c|}
\hline
Time & Task & Machine & Job \\
\hline
0-3 & Job1-Op1 & MachineA & Job1 \\
3-5 & Job2-Op1 & MachineA & Job2 \textcolor{red}{\textbf{(Error during repair)}}\\
5-6 & Job3-Op2 & MachineA & Job3 \textcolor{red}{\textbf{(Error during repair)}}\\
0-4 & Job3-Op1 & MachineB & Job3 \\
4-6 & Job1-Op2 & MachineB & Job1 \\
5-6 & Job2-Op2 & MachineC & Job2 \\
6-8 & Job1-Op3 & MachineC & Job1 \\
6-10 & Job2-Op3 & MachineB & Job2 \\
8-11 & Job3-Op3 & MachineC & Job3 \\
\hline
\end{tabular}
\end{table}

\subsubsection{Gemini 2.5 Pro Plan (Incorrect)}
\begin{table}[htbp]
\centering
\caption{Gemini Correct Plan}
\label{tab:gemini_correct}
\footnotesize
\begin{tabular}{|c|c|c|c|}
\hline
Time & Task & Machine & Job \\
\hline
0-3 & Job1-Op1 & MachineA & Job1 \\
0-4 & Job3-Op1 & MachineB & Job3 \\
3-5 & Job2-Op1 & MachineA & Job2 \textcolor{red}{\textbf{(Error during repair)}}\\
4-6 & Job1-Op2 & MachineB & Job1 \\
5-6 & Job2-Op2 & MachineC & Job2 \\
5-6 & Job3-Op2 & MachineA & Job3 \textcolor{red}{\textbf{(Error during repair)}}\\
6-8 & Job1-Op3 & MachineC & Job1 \\
6-10 & Job2-Op3 & MachineB & Job2 \\
8-11 & Job3-Op3 & MachineC & Job3 \\
\hline
\end{tabular}
\end{table}

\begin{figure}[htbp]
\centering
\begin{tikzpicture}[xscale=0.4, yscale=0.8]
\foreach \x in {0,...,15} {
  \draw[gray!60] (\x,0.5) -- (\x,3.5);
  \node[below] at (\x, 0.5) {\tiny \x};
}
\node[anchor=east] at (0,3) {MachineA};
\draw[gray!30, thick, ->] (0,3) -- (15,3);
\fill[gray!30] (4,2.6) rectangle (6,3.4);
\node[anchor=east] at (0,2) {MachineB};
\draw[gray!30, thick, ->] (0,2) -- (15,2);
\node[anchor=east] at (0,1) {MachineC};
\draw[gray!30, thick, ->] (0,1) -- (15,1);
\draw[fill=blue!30, draw=black] (0,2.7) rectangle (3,3.3) node[pos=.5]{\scriptsize Job1-Op1};
\draw[fill=red!30, draw=black] (0,1.7) rectangle (4,2.3) node[pos=.5]{\scriptsize Job3-Op1};
\draw[fill=red!70, draw=black] (3,2.7) rectangle (4,3.3) node[pos=.5]{\scriptsize Job3-Op2};
\draw[fill=blue!30, draw=black] (4,1.7) rectangle (6,2.3) node[pos=.5]{\scriptsize Job1-Op2};
\draw[fill=red!30, draw=black] (4,0.7) rectangle (7,1.3) node[pos=.5]{\scriptsize Job3-Op3};
\draw[fill=green!30, draw=black] (6,2.7) rectangle (8,3.3) node[pos=.5]{\scriptsize Job2-Op1};
\draw[fill=blue!30, draw=black] (7,0.7) rectangle (9,1.3) node[pos=.5]{\scriptsize Job1-Op3};
\draw[fill=green!30, draw=black] (9,0.7) rectangle (10,1.3) node[pos=.5]{\scriptsize Job2-Op2};
\draw[fill=green!30, draw=black] (10,1.7) rectangle (14,2.3) node[pos=.5]{\scriptsize Job2-Op3};
\end{tikzpicture}
\caption{Gantt Chart: GPT-4o (Incorrect)}
\label{fig:gpt-4o_(incorrect)}
\end{figure}

\begin{figure}[htbp]
\centering
\begin{tikzpicture}[xscale=0.4, yscale=0.8]
\foreach \x in {0,...,15} {
  \draw[gray!60] (\x,0.5) -- (\x,3.5);
  \node[below] at (\x, 0.5) {\tiny \x};
}
\node[anchor=east] at (0,3) {MachineA};
\draw[gray!30, thick, ->] (0,3) -- (15,3);
\fill[gray!30] (4,2.6) rectangle (6,3.4);
\node[anchor=east] at (0,2) {MachineB};
\draw[gray!30, thick, ->] (0,2) -- (15,2);
\node[anchor=east] at (0,1) {MachineC};
\draw[gray!30, thick, ->] (0,1) -- (15,1);
\draw[fill=blue!30, draw=black] (0,2.7) rectangle (3,3.3) node[pos=.5]{\scriptsize Job1-Op1};
\draw[fill=red!30, draw=black] (0,1.7) rectangle (4,2.3) node[pos=.5]{\scriptsize Job3-Op1};
\draw[fill=red!70, draw=black] (3,2.7) rectangle (5,3.3) node[pos=.5]{\scriptsize Job2-Op1};
\draw[fill=blue!30, draw=black] (4,1.7) rectangle (6,2.3) node[pos=.5]{\scriptsize Job1-Op2};
\draw[fill=green!30, draw=black] (5,0.7) rectangle (6,1.3) node[pos=.5]{\scriptsize Job2-Op2};
\draw[fill=red!30, draw=black] (6,2.7) rectangle (7,3.3) node[pos=.5]{\scriptsize Job3-Op2};
\draw[fill=blue!30, draw=black] (6,0.7) rectangle (8,1.3) node[pos=.5]{\scriptsize Job1-Op3};
\draw[fill=green!30, draw=black] (6,1.7) rectangle (10,2.3) node[pos=.5]{\scriptsize Job2-Op3};
\draw[fill=red!30, draw=black] (8,0.7) rectangle (11,1.3) node[pos=.5]{\scriptsize Job3-Op3};
\end{tikzpicture}
\caption{Gantt Chart: Claude 3.7 (Incorrect)}
\label{fig:claude_3.7_(incorrect)}
\end{figure}

\begin{figure}[htbp]
\centering
\begin{tikzpicture}[xscale=0.4, yscale=0.8]
\foreach \x in {0,...,15} {
  \draw[gray!60] (\x,0.5) -- (\x,3.5);
  \node[below] at (\x, 0.5) {\tiny \x};
}
\node[anchor=east] at (0,3) {MachineA};
\draw[gray!30, thick, ->] (0,3) -- (15,3);
\fill[gray!30] (4,2.6) rectangle (6,3.4);
\node[anchor=east] at (0,2) {MachineB};
\draw[gray!30, thick, ->] (0,2) -- (15,2);
\node[anchor=east] at (0,1) {MachineC};
\draw[gray!30, thick, ->] (0,1) -- (15,1);
\draw[fill=blue!30, draw=black] (0,2.7) rectangle (3,3.3) node[pos=.5]{\scriptsize Job1-Op1};
\draw[fill=red!70, draw=black] (3,2.7) rectangle (5,3.3) node[pos=.5]{\scriptsize Job2-Op1};
\draw[fill=red!70, draw=black] (5,2.7) rectangle (6,3.3) node[pos=.5]{\scriptsize Job3-Op2};
\draw[fill=red!30, draw=black] (0,1.7) rectangle (4,2.3) node[pos=.5]{\scriptsize Job3-Op1};
\draw[fill=blue!30, draw=black] (4,1.7) rectangle (6,2.3) node[pos=.5]{\scriptsize Job1-Op2};
\draw[fill=green!30, draw=black] (5,0.7) rectangle (6,1.3) node[pos=.5]{\scriptsize Job2-Op2};
\draw[fill=blue!30, draw=black] (6,0.7) rectangle (8,1.3) node[pos=.5]{\scriptsize Job1-Op3};
\draw[fill=green!30, draw=black] (6,1.7) rectangle (10,2.3) node[pos=.5]{\scriptsize Job2-Op3};
\draw[fill=red!30, draw=black] (8,0.7) rectangle (11,1.3) node[pos=.5]{\scriptsize Job3-Op3};
\end{tikzpicture}
\caption{Gantt Chart: DeepSeek R1 (Incorrect)}
\label{fig:deepseek_r1_(correct)}
\end{figure}

\begin{figure}[htbp]
\centering
\begin{tikzpicture}[xscale=0.4, yscale=0.8]
\foreach \x in {0,...,15} {
  \draw[gray!60] (\x,0.5) -- (\x,3.5);
  \node[below] at (\x, 0.5) {\tiny \x};
}
\node[anchor=east] at (0,3) {MachineA};
\draw[gray!30, thick, ->] (0,3) -- (15,3);
\fill[gray!30] (4,2.6) rectangle (6,3.4);
\node[anchor=east] at (0,2) {MachineB};
\draw[gray!30, thick, ->] (0,2) -- (15,2);
\node[anchor=east] at (0,1) {MachineC};
\draw[gray!30, thick, ->] (0,1) -- (15,1);
\draw[fill=blue!30, draw=black] (0,2.7) rectangle (3,3.3) node[pos=.5]{\scriptsize Job1-Op1};
\draw[fill=red!30, draw=black] (0,1.7) rectangle (4,2.3) node[pos=.5]{\scriptsize Job3-Op1};
\draw[fill=green!30, draw=black] (3,2.7) rectangle (5,3.3) node[pos=.5]{\scriptsize Job2-Op1};
\draw[fill=blue!30, draw=black] (4,1.7) rectangle (6,2.3) node[pos=.5]{\scriptsize Job1-Op2};
\draw[fill=green!30, draw=black] (5,0.7) rectangle (6,1.3) node[pos=.5]{\scriptsize Job2-Op2};
\draw[fill=red!30, draw=black] (5,2.7) rectangle (6,3.3) node[pos=.5]{\scriptsize Job3-Op2};
\draw[fill=blue!30, draw=black] (6,0.7) rectangle (8,1.3) node[pos=.5]{\scriptsize Job1-Op3};
\draw[fill=green!30, draw=black] (6,1.7) rectangle (10,2.3) node[pos=.5]{\scriptsize Job2-Op3};
\draw[fill=red!30, draw=black] (8,0.7) rectangle (11,1.3) node[pos=.5]{\scriptsize Job3-Op3};
\end{tikzpicture}
\caption{Gantt Chart: Gemini 2.5 Pro (Correct)}
\label{fig:gemini_2.5_pro_(correct)}
\end{figure}

\subsubsection{Summary of Results}

\begin{table}[htbp]
\centering
\caption{Summary of Model Performances (Pass@1)}
\label{tab:summary_models}
\footnotesize
\resizebox{0.5\textwidth}{!}{
\begin{tabular}{|l|c|c|c|c|}
\hline
\textbf{Model} & \textbf{Validation Rate } & \textbf{Correct Steps} & \textbf{Machine Repair} & \textbf{Job Order} \\
\textbf{} & \textbf{(Pass@1)} & \textbf{} & \textbf{Errors} & \textbf{Errors} \\
\hline
GPT-4o & 1/10 & 1 & 8 & 1 \\
Claude 3.7 & 2/10 & 2 & 6 & 2 \\
DeepSeek R1 & 7/10 & 7 & 0 & 3 \\
Gemini 2.5 Pro & -- & \textit{-} & \textit{-} & \textit{-} \\
\hline
\end{tabular}
}
\end{table}

\subsection{Experiment 2: Pass@2}

\subsubsection{GPT-4o (1st Incorrect, 2nd Correct)}
\begin{itemize}
    \item Incorrect plan is generated due to constraint violation of ``order'' or ``machine repair'' at first prompt.
    \item Correct plan is generated in second prompt.
\end{itemize}

\begin{table}[htbp]
\centering
\caption{Final GPT-4o Plan (Correct)}
\label{tab:correct_reactive_schedule}
\footnotesize
\begin{tabular}{|c|c|c|c|}
\hline
Time & Task & Machine & Job \\
\hline
0–3   & Job1-Op1 & MachineA & Job1 \\
3–5   & Job1-Op2 & MachineB & Job1 \\
5–7   & Job1-Op3 & MachineC & Job1 \\
0–4   & Job3-Op1 & MachineB & Job3 \\
6–7   & Job3-Op2 & MachineA & Job3 \\
7–10  & Job3-Op3 & MachineC & Job3 \\
10–12 & Job2-Op1 & MachineA & Job2 \\
12–13 & Job2-Op2 & MachineC & Job2 \\
13–17 & Job2-Op3 & MachineB & Job2 \\
\hline
\end{tabular}
\end{table}

        
        




\subsubsection{Claude 3.7 (1st Incorrect, 2nd Correct)}
\begin{itemize}
    \item Incorrect plan is generated due to constraint violation of ``job separation'' and ``machine repair'' at first prompt.
    \item Correct plan is generated in second prompt.
\end{itemize}

\begin{table}[htbp]
\centering
\caption{Final Claude 3.7 Plan (Correct)}
\label{tab:Claude_correct_reactive_schedule}
\footnotesize
\begin{tabular}{|c|c|c|c|}
\hline
Time & Task & Machine & Job \\
\hline
0–2   & Job2-Op1 & MachineA & Job2 \\
0–4   & Job3-Op1 & MachineB & Job3 \\
2–3   & Job2-Op2 & MachineC & Job2 \\
3–7   & Job2-Op3 & MachineB & Job2 \\
6–9   & Job1-Op1 & MachineA & Job1 \\
9–10  & Job3-Op2 & MachineA & Job3 \\
9–11  & Job1-Op2 & MachineB & Job1 \\
10–13 & Job3-Op3 & MachineC & Job3 \\
11–13 & Job1-Op3 & MachineC & Job1 \\
\hline
\end{tabular}
\end{table}

\subsubsection{DeepSeekR1 (1st Incorrect, 2nd Correct)}
\begin{itemize}
    \item Incorrect plan is generated due to constraint violation of ``order'' or ``machine repair'' at first prompt.
    \item Correct plan is generated in second prompt.
\end{itemize}

\begin{table}[htbp]
\centering
\caption{Final DeepSeek R1-Plan (Correct)}
\label{tab:deepseek_contiguous_schedule}
\footnotesize
\begin{tabular}{|c|c|c|c|}
\hline
Time & Task & Machine & Job \\
\hline
0--3   & Job1-Op1 & MachineA & Job1 \\
0--4   & Job3-Op1 & MachineB & Job3 \\
4--6   & \textit{Idle} & MachineA & -- \\
4--6   & Job1-Op2 & MachineB & Job1 \\
5--6   & Job2-Op2 & MachineC & Job2 \\
6--8   & Job1-Op3 & MachineC & Job1 \\
6--8   & Job2-Op1 & MachineA & Job2 \\
8--9   & Job3-Op2 & MachineA & Job3 \\
8--9   & Job2-Op2 & MachineC & Job2 \\
9--12  & Job3-Op3 & MachineC & Job3 \\
9--13  & Job2-Op3 & MachineB & Job2 \\
\hline
\end{tabular}
\end{table}

\subsubsection{Gemini 2.5 Pro (-)}
\begin{itemize}
    \item Incorrect plan is generated due to constraint violation of ``order'' or ``machine repair'' at first prompt.
    \item Correct plan is generated in second prompt.
\end{itemize}

\subsection{Summary of Results}

\begin{table}[htbp]
\centering
\caption{Summary of Model Performances on One Instance (Pass@2)}
\label{tab:summary_models_detailed}
\footnotesize
\resizebox{0.5\textwidth}{!}{
\begin{tabular}{|l|c|c|c|}
\hline
\textbf{Model} & \textbf{Correctness} & \textbf{Makespan / Optimality} & \textbf{Time Usage} \\
\hline
GPT-4o & \textcolor{green}{Correct} & Suboptimal & $\sim$10 sec / prompt \\
Claude 3.7 & \textcolor{green}{Correct} & Optimal & $\sim$10 sec / prompt \\
DeepSeek R1 & \textcolor{green}{Correct} & Optimal & $>$5 min / prompt \\
Gemini 2.5 Pro & -- & -- & -- \\
\hline
\end{tabular}
}
\end{table}

\subsection{Meta Planner Algorithm 1 - Mathematical Formulation (MIP Model)}

Let $s_{ij}$ be the start time of operation $j$ in job $i$, and $p_{ij}$ be its processing time.

\textbf{Objective:}
\begin{equation}
\text{Minimize} \quad \max_{i,j} (s_{ij} + p_{ij})
\end{equation}

\textbf{Subject to:}
\begin{itemize}
    \item \textbf{Precedence Constraint:} $s_{ij+1} \geq s_{ij} + p_{ij}$
    \item \textbf{Machine Conflict Constraint:} For operations $a$ and $b$ on the same machine:
    \begin{equation}
        (s_{a} \geq s_{b} + p_{b}) \quad \text{or} \quad (s_{b} \geq s_{a} + p_{a})
    \end{equation}
    \item \textbf{Non-overlapping Steps Constraint (Prompt 2):} For  operations within same job on different machines:
    \begin{equation}
        s_{ij+1} \geq s_{ij} + p_{ij}
    \end{equation}
\end{itemize}

\noindent Where variables are non-negative.

\subsection{Error Analysis and Conclusion}

This case study highlights the complexity of job shop scheduling under dynamic and adversarial conditions, such as machine failures. 1) Models like DeepSeek and Gemini demonstrated strong performance in adapting to changes and producing valid reactive plans. 2) GPT-4o showed some improvements under tighter constraints but was less optimal, while Claude 3.7 struggled with adapting to both simple and strict scheduling constraints.

Structured meta-planning ($\mathcal{MP}$) with explicit constraint modeling significantly improves robustness, demonstrating the necessity of careful system design and constraint tracking in real-world multi-agent or multi-task optimization scenarios.

Future extensions could involve 1) adding stochastic delays (will introduce in next section), 2) different sub-meta planner algorithms, such as resource contention models, 3) different objectives such as First In First Out (FIFO), Dynamic Programming, 4) and complex, multi-objective trade-offs beyond makespan, such as energy cost or fairness in task allocations.

\vspace{0.2in}

\newpage
\subsection{Leaderboards}
\begin{table}[htbp]
\centering
\caption{Leaderboard 1: ALAS-Static vs. ALAS-Dynamic Makespan Plan Generation Optimality Summary (ALAS-static is better on DMU)}
\label{tab:llm_vs_ALAS_summary_gap1}
\resizebox{0.5\textwidth}{!}{
\begin{tabular}{llrrrrrrrrrrrrrr}
\toprule
\textbf{Dataset} & \textbf{Size} & \textbf{Random} & \textbf{LPT} & \textbf{SPT} & \textbf{STPT} & \textbf{MPSR} & \textbf{DRL-Liu} & \textbf{GP} \\
\midrule
DMU03 & 20 × 15 & 3827 & 4592 & 3630 & 4232 & 3435 & 3303 & 3540 \\
DMU04 & 20 × 15 & 3889 & 4047 & 3541 & 4642 & 3355 & 3321 & 3406 \\
DMU08 & 20 × 20 & 4228 & 4551 & 4714 & 4459 & 3999 & 4098 & 3802 \\
DMU09 & 20 × 20 & 4094 & 4511 & 4283 & 4690 & 3869 & 3753 & 4196 \\
DMU13 & 30 × 15 & 5451 & 5580 & 4813 & 5207 & 4759 & 4708 & 4765 \\
DMU14 & 30 × 15 & 5306 & 5591 & 4583 & 4811 & 4238 & 4124 & 4289 \\
DMU18 & 30 × 20 & 5326 & 5810 & 6231 & 5480 & 5003 & 4800 & 4696 \\
DMU19 & 30 × 20 & 5174 & 5787 & 5126 & 5203 & 4930 & 4837 & 4666 \\
DMU23 & 40 × 15 & 5948 & 7045 & 6250 & 6521 & 5383 & 5240 & 5391 \\
DMU24 & 40 × 15 & 6078 & 6484 & 5503 & 6595 & 5358 & 5319 & 5560 \\
DMU28 & 40 × 20 & 6737 & 7322 & 6558 & 7697 & 5927 & 5948 & 6017 \\
DMU29 & 40 × 20 & 6602 & 7386 & 6565 & 7690 & 6107 & 5824 & 6236\\
DMU33 & 50 × 15 & 6890 & 8779 & 7361 & 7631 & 6282 & 6458 & 6109 \\
DMU34 & 50 × 15 & 7523 & 7991 & 7026 & 7740 & 6359 & 6284 & 6327 \\
DMU38 & 50 × 20 & 7685 & 9051 & 7954 & 8555 & 7604 & 7275 & 7267 \\
DMU39 & 50 × 20 & 8097 & 8514 & 7592 & 8908 & 6953 & 6776 & 6941 \\
Mean & NaN & 5803.44 & 6440.06 & 5733.12 & 6253.81 & 5222.56 & 5129.25 & 5200.50 \\
\midrule
\textbf{Gap to UB (\%)} & -- & 37.28 & 52.34 & 35.62 & 47.93 & 23.54 & 21.33 & 23.02 \\
\bottomrule
\end{tabular}
}
\end{table}

\begin{table}[htbp]
\centering
\caption{Leaderboard 1 (cont): ALAS-Static vs. ALAS-Dynamic Makespan Plan Generation Optimality Summary (ALAS-static is better on DMU)}
\label{tab:llm_vs_ALAS_summary_gap2}
\resizebox{0.5\textwidth}{!}{
\begin{tabular}{llrrrrrrrrrrrrrr}
\toprule
\textbf{Dataset} & \textbf{Size} & \textbf{GEP} & \textbf{SeEvo} & \textbf{SeEvo} & \textbf{UB} & \textbf{ALAS-dynamic} & \textbf{ALAS-static}\\
 &  &  & \textbf{(GLM3)} & \textbf{(GPT3.5)} & & \textbf{(Claude-3.7 heuristics)} & \textbf{(Claude-3.7 heuristics)} \\

\midrule
DMU03 & 20 × 15 & 3651 & 3462 & 3238 & \textbf{2731} & 3356 & \textbf{3462} \\
DMU04 & 20 × 15 & 3499 & 3235 & 3212 & \textbf{2669} & 3352 & \textbf{3235} \\
DMU08 & 20 × 20 & 4023 & 3728 & 3728 & \textbf{3188} & 3906 & \textbf{3728} \\
DMU09 & 20 × 20 & 4136 & 3857 & 3828 & \textbf{3092} & 3731 & \textbf{3857} \\
DMU13 & 30 × 15 & 4812 & 4658 & 4709 & \textbf{3681} & 4524 & \textbf{4658} \\
DMU14 & 30 × 15 & 4213 & 3980 & 3980 & \textbf{3394} & 4195 & \textbf{3980} \\
DMU18 & 30 × 20 & 4917 & 4724 & 4724 & \textbf{3844} & 4675 & \textbf{4724} \\
DMU19 & 30 × 20  & 5245 & 4715 & 4816 & \textbf{3768} & 4774 & \textbf{4715} \\
DMU23 & 40 × 15 & 5595 & 5151 & 5258 & \textbf{4668} & 5805 & \textbf{5151} \\
DMU24 & 40 × 15 & 5458 & 5226 & 5316 & \textbf{4648} & 5750 & \textbf{5226} \\
DMU28 & 40 × 20 & 6142 & 5838 & 5944 & \textbf{4692} & 5550 & \textbf{5838} \\
DMU29 & 40 × 20 & 6224 & 5941 & 5825 & \textbf{4691} & 5661 & \textbf{5941} \\
DMU33 & 50 × 15 & 6081 & 6029 & 6029 & \textbf{5728} & 7158 & \textbf{6029} \\
DMU34 & 50 × 15 & 6279 & 6148 & 6146 & \textbf{5385} & 6597 & \textbf{6148} \\
DMU38 & 50 × 20 & 7501 & 7168 & 7170 & \textbf{5713} & 7119 & \textbf{7168} \\
DMU39 & 50 × 20 & 7124 & 6693 & 6590 & \textbf{5747} & 6799 & \textbf{6693} \\
Mean & NaN & 5306.25 & 5034.56 & 5032.06 & \textbf{4227.44} & 5184.50 & \textbf{5034.56} \\
\midrule
\textbf{Gap to UB (\%)} & -- & 25.52 & 19.09 & 19.03 & -- & 22.74 & \textbf{19.09} \\
\bottomrule
\end{tabular}
}
\end{table}

\begin{table}[htbp]
\centering
\caption{Leaderboard 2: ALAS-Static vs. ALAS-Dynamic Makespan Plan Generation Optimality Summary (ALAS-Dynamic is better on TA)}
\label{tab:llm_vs_ALAS_summary_gap3}
\resizebox{0.5\textwidth}{!}{
\begin{tabular}{llrrrrrrrrrrrrr}
\toprule
\textbf{Dataset} & \textbf{Size} & \textbf{LSO} & \textbf{SPT/TWKR} & \textbf{DRL-Chen} & \textbf{DRL-Zhang} & \textbf{DRL-Liu} & \textbf{GP} \\
\midrule
TA01 & 15 × 15 & 1957 & 1664 & 1711 & 1433 & 1492 & 1547 \\
TA02 & 15 × 15 & 1759 & 1538 & 1639 & 1544 & 1425 & 1565 \\
TA51 & 50 × 15 & 3844 & 3768 & 3762 & 3599 & 3608 & 3603 \\
TA52 & 50 × 15 & 3715 & 3588 & 3511 & 3341 & 3524 & 3346 \\
TA61 & 50 × 20 & 4188 & 3752 & 3633 & 3654 & 3548 & 3685 \\
TA71 & 100 × 20 & 6754 & 6705 & 6321 & 6452 & 6289 & 6305 \\
TA72 & 100 × 20 & 6674 & 6351 & 6232 & 5695 & 6002 & 5776 \\
\midrule
\textbf{Mean} & - & 4127.29 & 3909.43 & 3829.86 & 3674 & 3698.29 & 3689.57 \\
\textbf{Gap to UB (\%)} & -- 
& 34.31 & 27.23 & 24.66 & 18.48 & 19.39 
& 20.12 \\
\bottomrule
\end{tabular}
}
\end{table}

\begin{table}[htbp]
\centering
\caption{Leaderboard 2 (cont): ALAS-Static vs. ALAS-Dynamic Makespan Plan Generation Optimality Summary (ALAS-Dynamic is better on TA)}
\label{tab:llm_vs_ALAS_summary_gap4}
\resizebox{0.5\textwidth}{!}{
\begin{tabular}{llrrrrrrrrrrrrr}
\toprule
\textbf{Dataset} & \textbf{Size} & \textbf{GEP} & \textbf{SeEvo} & \textbf{SeEvo} & \textbf{UB} & \textbf{ALAS-dynamic} & \textbf{ALAS-static}\\
 &  &  & \textbf{(GLM3)} & \textbf{(GPT3.5)} & & \textbf{(Claude-3.7 heuristics)} & \textbf{(Claude-3.7 heuristics)} \\
 
\midrule
TA01 & 15 × 15 & 1547 & 1427 & 1427 & \textbf{1231} & \textbf{1243} & 1231 \\
TA02 & 15 × 15 & 1486 & 1465 & 1437 & \textbf{1244} & \textbf{1252} & 1244 \\
TA51 & 50 × 15 & 3668 & 3364 & 3412 & \textbf{2760} & \textbf{2766}& 2760 \\
TA52 & 50 × 15 & 3324 & 3286 & 3245 & \textbf{2756} & \textbf{2819} & 2756 \\
TA61 & 50 × 20 & 3642 & 3529 & 3537 & \textbf{2868} & \textbf{2905} & F \\
TA71 & 100 × 20 & 6278 & 6071 & 6099 & \textbf{5464} & \textbf{5478} & 5464 \\
TA72 & 100 × 20 & 5625 & 5604 & 5575 & \textbf{5181} & \textbf{5198} & F \\
\midrule
Mean & -- & 3652.86 & 3535.14 & 3533.14 & \textbf{3072} & \textbf{3094.43} & -- \\
\textbf{Gap to UB (\%)} & -- & 18.91 & 15.10 & 14.99 & -- 
& \textbf{0.86} & -- \\
\bottomrule
\end{tabular}
}
\end{table}

\begin{table}[htbp]
\centering
\caption{Leaderboard 3: ALAS-Static vs. ALAS-Dynamic Makespan Plan Generation Validity and Optimality on Additional Benchmark Instances (ABZ, SWV, and YN)}
\label{tab:static_dynamic_gap}
\resizebox{0.5\textwidth}{!}{
\begin{tabular}{lcccccccccc}
\toprule
\textbf{Dataset} & \textbf{Size} & \textbf{UB} & \textbf{Static Makespan} & \textbf{Valid Static} & \textbf{Dynamic Min} & \textbf{Dynamic Max}\\
\midrule
abz07 & 20×15 & \textbf{656} & 656 & True & 659 & 978 \\
abz08 & 20×15 & \textbf{667} & 667 & True & 701 & 983 \\
abz09 & 20×15 & \textbf{678} & 678 & True & 679 & 975 \\
swv01 & 20×10 & \textbf{1407} & 1406 & - & 1429 & 2100 \\
swv02 & 20×10 & \textbf{1475} & 1475 & True & 1481 & 2177\\
swv03 & 20×10 & \textbf{1398} & 1398 & True & 1429 & 2073 \\
swv04 & 20×10 & \textbf{1464} & 1464 & True & 1466 & 2168 \\
swv05 & 20×10 & \textbf{1424} & 1424 & True & 1430 & 2086 \\
swv06 & 20×15 & \textbf{1667} & 1667 & True & 1716 & 2485 \\
swv07 & 20×15 & \textbf{1595} & 1595 & True & 1621 & 2388 \\
swv08 & 20×15 & \textbf{1751} & 1751 & True & 1774 & 2535 \\
swv09 & 20×15 & \textbf{1655} & 1655 & True & 1672 & 2446 \\
swv10 & 20×15 & \textbf{1743} & 1743 & True & 1817 & 2603 \\
swv11 & 50×10 & \textbf{2983} & 2983 & True & 3099 & 4470 \\
swv12 & 50×10 & \textbf{2972} & 2972 & True & 2992 & 4423 \\
swv13 & 50×10 & \textbf{3104} & 3104 & True & 3144 & 4573 \\
swv14 & 50×10 & \textbf{2968} & 2968 & True & 2981 & 4396 \\
swv15 & 50×10 & \textbf{2885} & 2885 & True & 2912 & 4301 \\
yn01 & 20×20 & \textbf{884} & 884 & True & 888 & 1293 \\
yn02 & 20×20 & \textbf{904} & 904 & True & 942 & 1321 \\
yn03 & 20×20 & \textbf{892} & 892 & True & 900 & 1320 \\
yn04 & 20×20 & \textbf{968} & 968 & True & 980 & 1450 \\
\midrule
\textbf{Mean} & -- & \textbf{1662.87} & 1662.74 & -- & 1685.00 & 2483.61 \\
\textbf{Gap to UB (\%)} & -- & -- & -- & -- & -- & -- \\

\bottomrule
\end{tabular}
}
\end{table}

\begin{table}[htbp]
\centering
\caption{Leaderboard 3 (cont): ALAS-Static vs. ALAS-Dynamic Makespan Plan Generation Validity and Optimality on Additional Benchmark Instances (ABZ, SWV, and YN)}
\label{tab:static_dynamic_gap2}
\resizebox{0.5\textwidth}{!}{
\begin{tabular}{lcccccccccc}
\toprule
\textbf{Dataset} & \textbf{Size} & \textbf{ALAS-Static} & \textbf{ALAS-Dynamic} & \textbf{Static Gap} & \textbf{Dynamic Gap} \\
 &  & \textbf{ Valid Rate} & \textbf{Valid Rate} & \textbf{ (\%)} & \textbf{(\%)} \\
\midrule
abz07 & 20×15 &  \textbf{1.0} & \textbf{1.0} & \textbf{0.00\%} & \textbf{0.46\% }\\
abz08 & 20×15  & \textbf{1.0} & \textbf{1.0} & \textbf{0.00\%} & \textbf{5.10\%} \\
abz09 & 20×15 & \textbf{1.0} & \textbf{1.0}& \textbf{0.00\%} & \textbf{0.15\%} \\
swv01 & 20×10 &  \textbf{--} & \textbf{1.0}  & \textbf{--} & \textbf{1.56\%} \\
swv02 & 20×10 & \textbf{1.0}  & \textbf{1.0}& \textbf{0.00\%} & \textbf{0.41\%} \\
swv03 & 20×10 & \textbf{1.0}  & \textbf{1.0}& \textbf{0.00\%} & \textbf{2.22\%}\\
swv04 & 20×10 & \textbf{1.0}  & \textbf{1.0}& \textbf{0.00\%} & \textbf{0.14\%} \\
swv05 & 20×10 & \textbf{1.0}  & \textbf{1.0}& \textbf{0.00\%} & \textbf{0.42\%} \\
swv06 & 20×15 & \textbf{1.0}  & \textbf{1.0}& \textbf{0.00\%} & \textbf{2.94\%} \\
swv07 & 20×15 & \textbf{1.0}  & \textbf{1.0}& \textbf{0.00\%} & \textbf{1.63\%} \\
swv08 & 20×15 & \textbf{1.0}  & \textbf{1.0}& \textbf{0.00\%} & \textbf{1.31\%} \\
swv09 & 20×15 & \textbf{1.0}  & \textbf{1.0}& \textbf{0.00\%} & \textbf{1.03\%} \\
swv10 & 20×15 & \textbf{1.0}  & \textbf{1.0}& \textbf{0.00\%} & \textbf{4.24\%} \\
swv11 & 50×10 & \textbf{1.0}  & \textbf{1.0}& \textbf{0.00\%} & \textbf{3.89\%} \\
swv12 & 50×10 & \textbf{1.0}  & \textbf{1.0}& \textbf{0.00\%} & \textbf{0.67\%} \\
swv13 & 50×10 & \textbf{1.0}  & \textbf{1.0}& \textbf{0.00\%} & \textbf{1.29\%} \\
swv14 & 50×10 & \textbf{1.0}  & \textbf{1.0}& \textbf{0.00\%} & \textbf{0.44\%} \\
swv15 & 50×10 & \textbf{1.0}  & \textbf{1.0}& \textbf{0.00\%} & \textbf{0.94\%} \\
yn01 & 20×20 & \textbf{1.0}  & \textbf{1.0}& \textbf{0.00\%} & \textbf{0.45\%} \\
yn02 & 20×20 & \textbf{1.0}  & \textbf{1.0}& \textbf{0.00\%} & \textbf{4.20\%} \\
yn03 & 20×20 & \textbf{1.0}  & \textbf{1.0}& \textbf{0.00\%} & \textbf{0.90\%} \\
yn04 & 20×20 & \textbf{1.0}  & \textbf{1.0}& \textbf{0.00\%} & \textbf{1.24\%} \\
\midrule
\textbf{Mean} & -- & \textbf{0.9545} & \textbf{\textbf{1.0}} & \textbf{--} & \textbf{--} \\
\textbf{Gap to UB (\%)} & -- & -- & -- & \textbf{--} & \textbf{1.65\%} \\

\bottomrule
\end{tabular}
}
\end{table}

\newpage
\subsection{Datasets}
\begin{figure}[b!]
    \centering
    \begin{subfigure}[b]{0.49\textwidth}
        \centering \includegraphics[width=\textwidth,height=\dimexpr 0.25\textwidth]{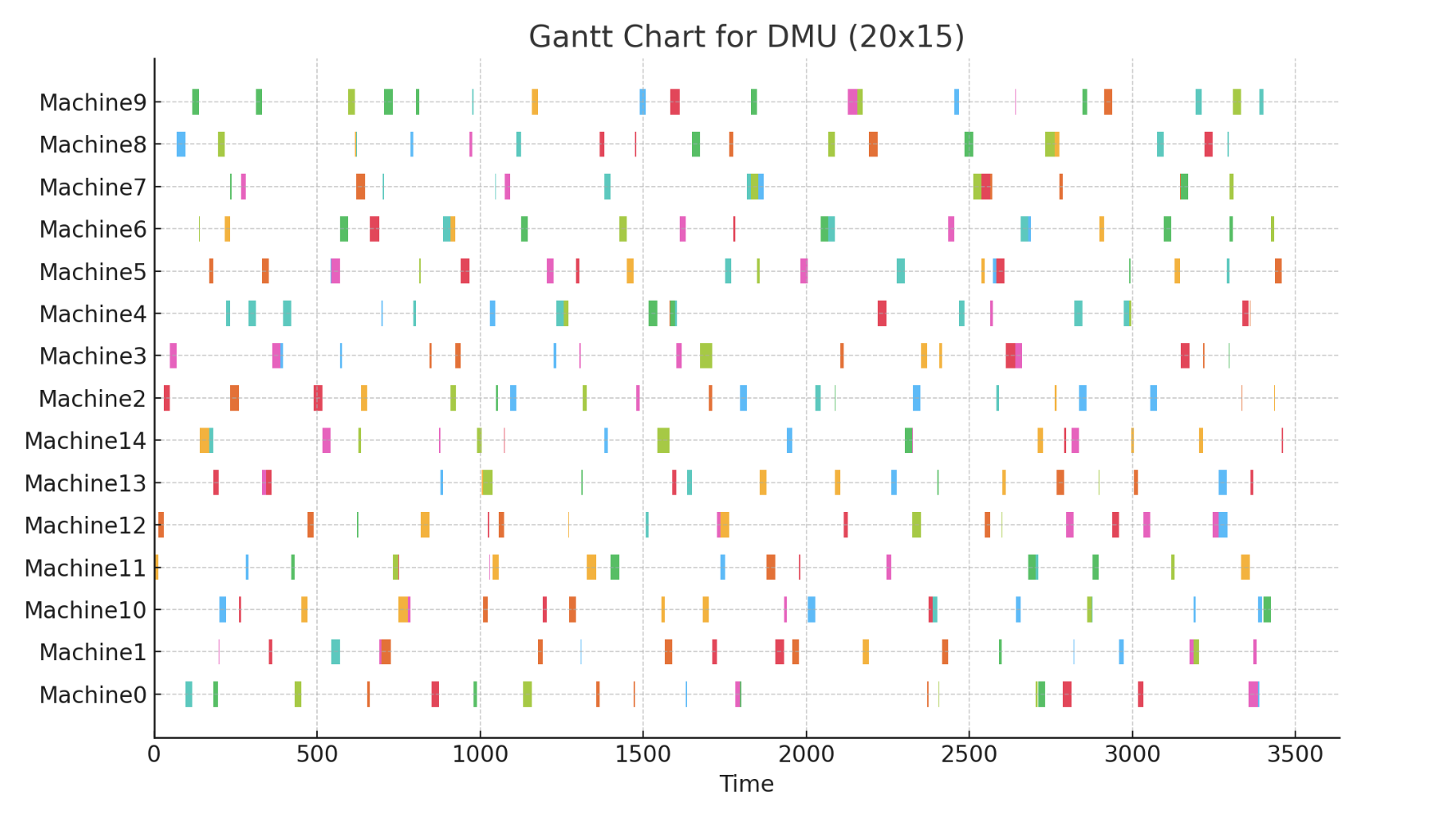}
        \caption{\textbf{rcmax\_20\_15\_5} (J=20, M=15)}
        \label{fig:gantt_rcmax}
    \end{subfigure}
    
    \hspace{-.2in}
    \begin{subfigure}[b]{0.49\textwidth}
        \centering
        \includegraphics[width=\textwidth,height=\dimexpr 0.25\textwidth]{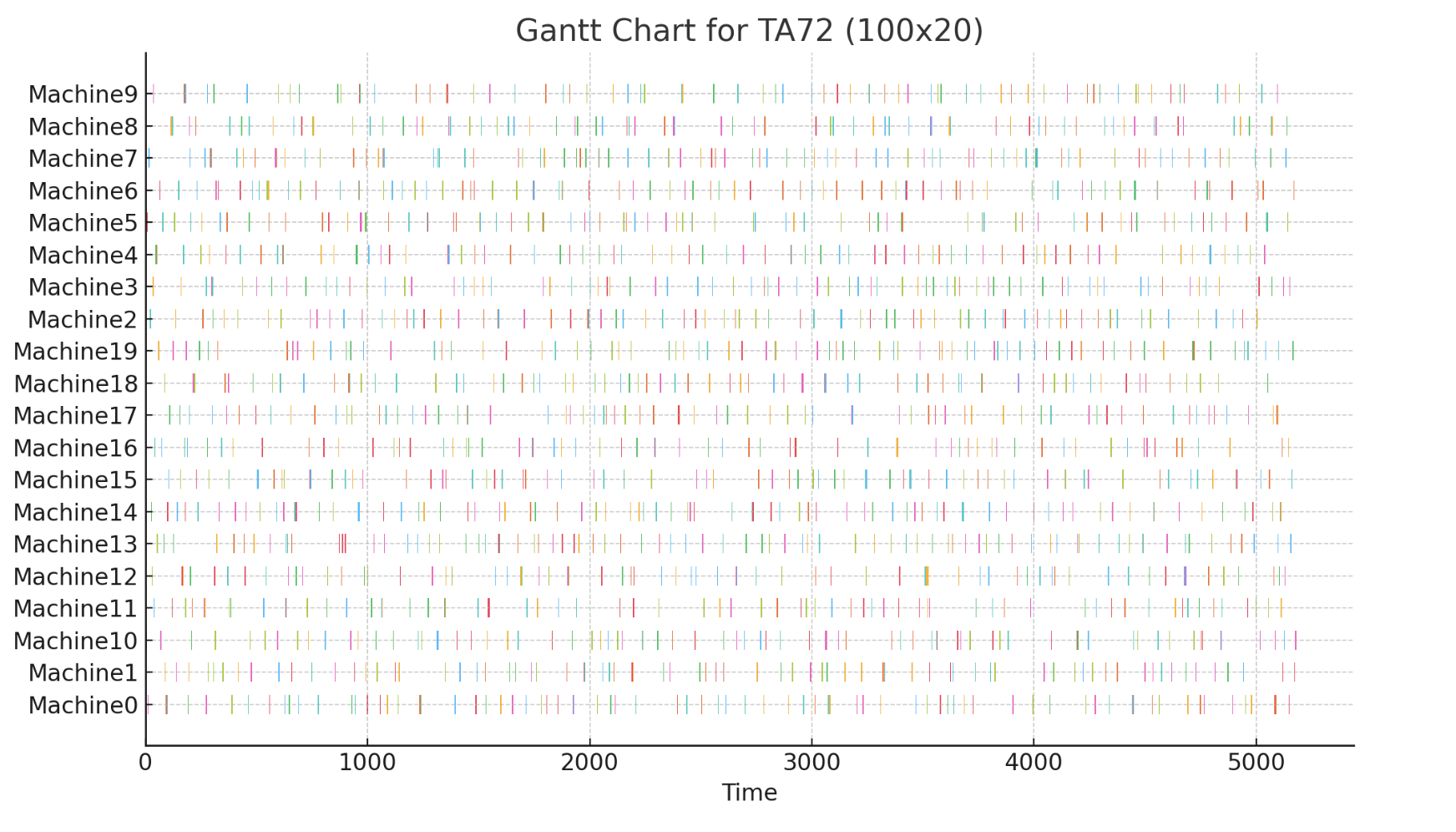}
        \caption{\textbf{Abz07} (J=20, M=15)}
        \label{fig:gantt_abz07}
    \end{subfigure}
    
    \vspace{0.3cm}
    
    \begin{subfigure}[b]{0.49\textwidth}
        \centering
        \includegraphics[width=\textwidth,height=\dimexpr 0.25\textwidth]{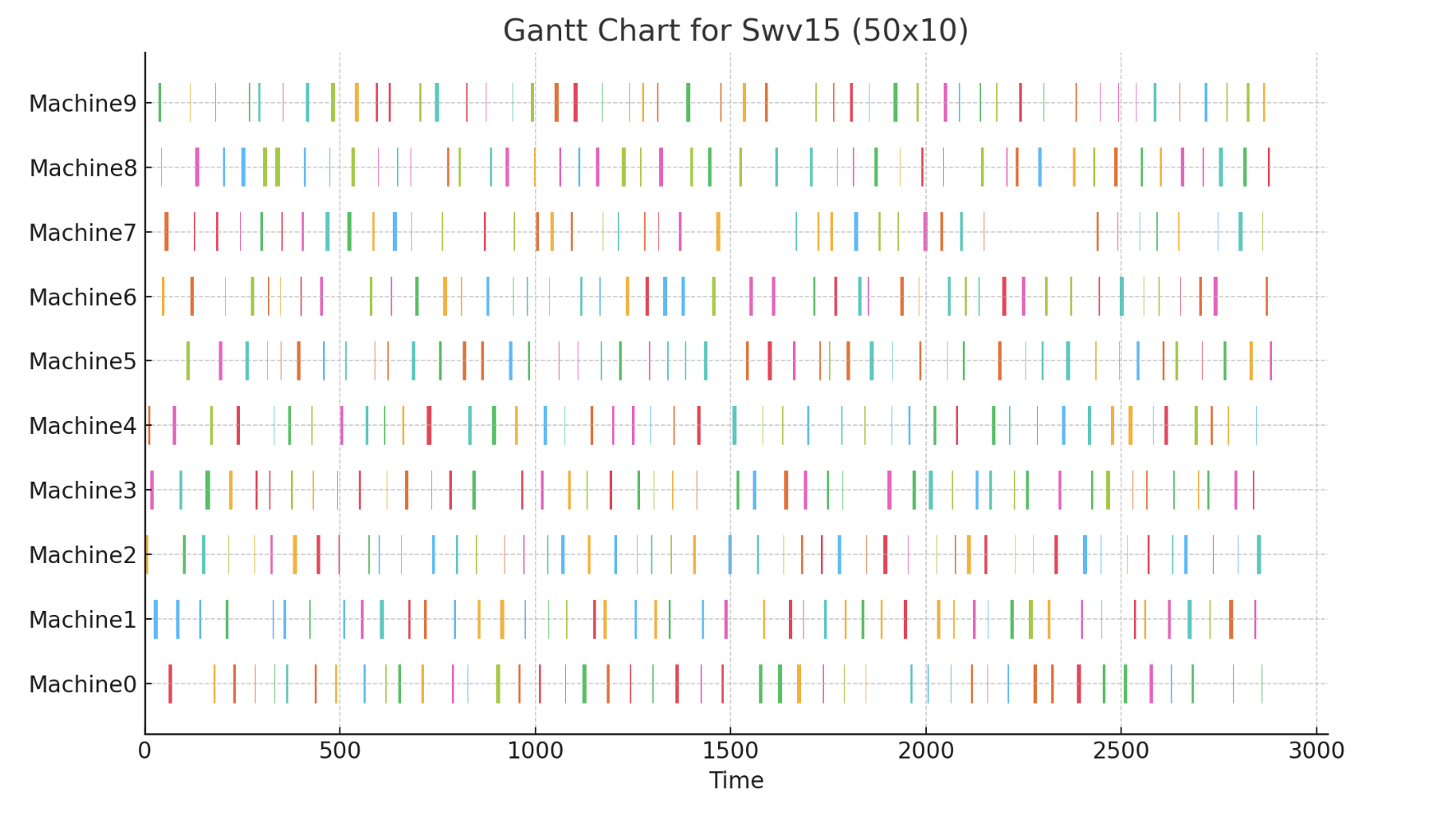}
        \caption{\textbf{Swv15} (J=50, M=10)}
        \label{fig:gantt_swv15}
    \end{subfigure}
    \hspace{-.2in}
    \begin{subfigure}[b]{0.49\textwidth}
        \centering
        \includegraphics[width=\textwidth,height=\dimexpr 0.25\textwidth]{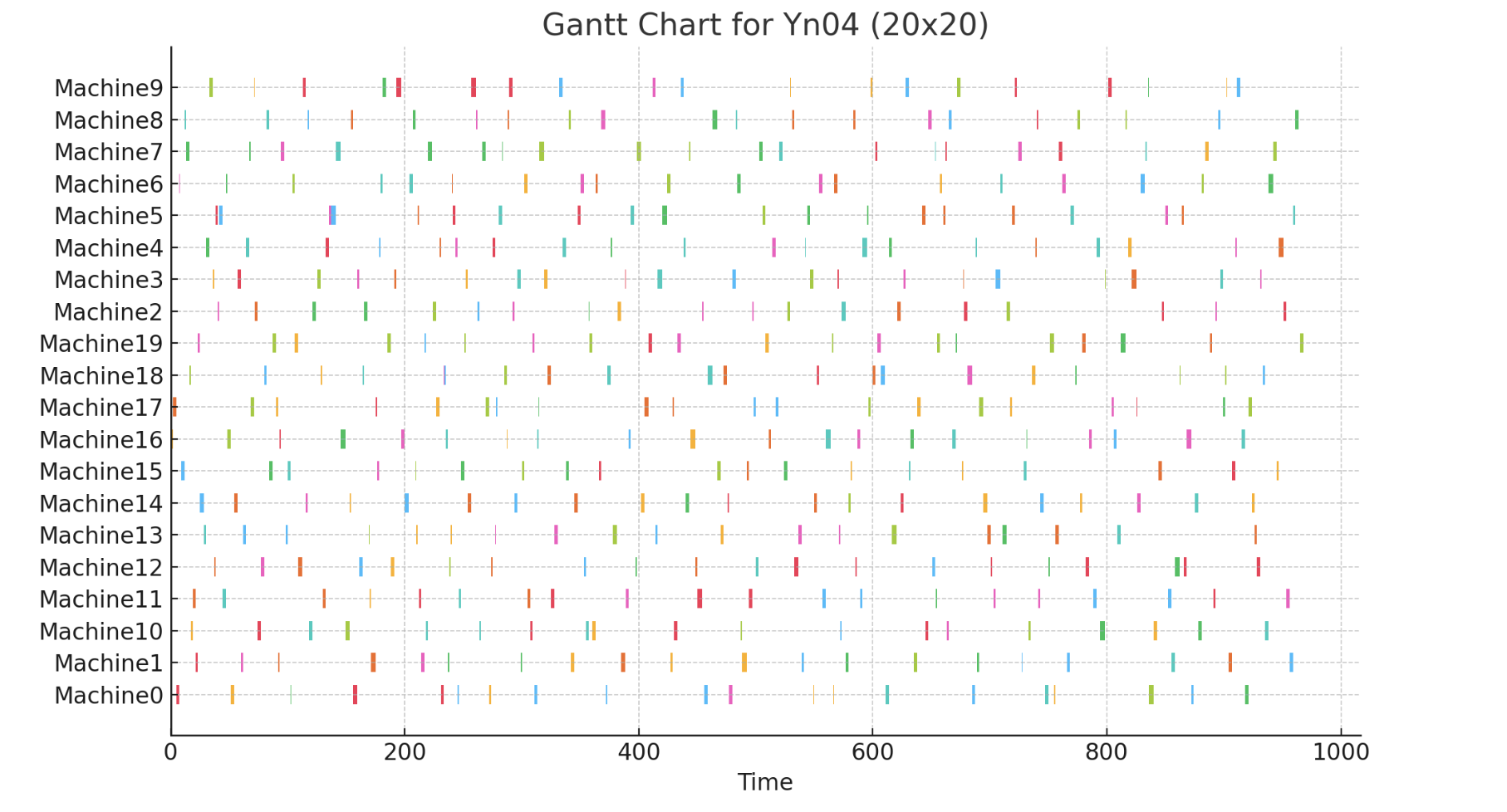}
        \caption{\textbf{Yn04} (J=20, M=20)}
        \label{fig:gantt_yn04}
    \end{subfigure}
    
    \caption{Gantt charts of optimized schedules produced by \texttt{ALAS} for four representative JSSP benchmark instances with varying job and machine counts. These visualizations demonstrate \texttt{ALAS}'s ability to efficiently allocate resources and minimize makespan across different problem scales. The larger instance TA72 (J=100, M=20) is available in the supplementary materials.}
    \label{fig:gantt_charts}
\end{figure}

We provide four examples from the following datasets in Figure~\ref{fig:gantt_charts}.
\begin{itemize}
    \item \textit{Demirkol-DMU}: Stress-tests scalability with job priority and routing heterogeneity across large instances (e.g., 20×15 to 50×20, in total 80 instances).
    \item \textit{Taillard (TA)}: Medium to large-scale JSSPs (15×15 to 100×20, in total 80 instances) with uniform mappings and tight constraints.
    \item \textit{Adams-Pinedo (ABZ)}: Smaller but tightly constrained problems (e.g., 20×15, in total 3 instances) for testing core optimization.
    \item \textit{Swv Benchmark (SWV)}: Dense overlap problem instances (20×10, 50×10, in total 15 instances) focusing on constraint satisfaction.
    \item \textit{Yamada–Nakano (YN)}: Realistic manufacturing settings (e.g., 20×20, in total 4 instances) evaluating machine utilization vs. flow time.
\end{itemize}

\newcommand{\ganttHeightScale}{0.5}
\newpage
\section{JSSP Problem Implementation (J1 and J2)}
\label{app:p11}

This appendix presents an implementation of \textbf{J1 and J2}: Job Shop Scheduling without and with interrupts, using LangGraph~\cite{langgraph2024}. We show JSSP-simple-static as an example, and we will release JSSP-simple-dynamic, JSSP-complex-static, and JSSP-complex-dynamic in future releases.

\subsection{Agentic Workflow Formulation}

In this first stage, we define agents to manage the nodes of the workflow, including local working job agent, and global supervision agent. At the end, we use the \texttt{>>} syntax to specify dependencies among agents. The transition from problem specifications to workflow formulation is handled automatically by MACI in LangGraph. The code also supports AutoGen, CrewAI, Swarm, and is extendable to other multi-agent framework.

\begin{lstlisting}[style=PythonStyle, caption=Datasets Preprocessing and Loading, label=lst:agent_pipeline4]
import sys
import os
from dotenv import load_dotenv
from openai import OpenAI
import google.generativeai as genai
from anthropic import Anthropic
# from deepseek import Deepseek
from datetime import datetime
import logging

# Set up project root and src path
dir_path = os.path.dirname(os.path.abspath(__file__))
project_root = os.path.abspath(os.path.join(dir_path, `..'))
sys.path.append(os.path.join(project_root, `src'))

from multi_agent.MAPLE import MAPLE
from multi_agent.agent import Agent

# Load environment variables
load_dotenv()

# Configure logging
def setup_logging(dataset):
    # Create results directory if it doesn't exist
    results_dir = ``./results''
    if not os.path.exists(results_dir):
        os.makedirs(results_dir)
        print(f``Created results directory: {results_dir}'')

    # Create a logger
    logger = logging.getLogger()
    logger.setLevel(logging.INFO)

    # Create file handler
    log_file = f``./results/{dataset}_dmu.txt''
    file_handler = logging.FileHandler(log_file)
    file_handler.setLevel(logging.INFO)

    # Create console handler
    console_handler = logging.StreamHandler()
    console_handler.setLevel(logging.INFO)

    # Create formatter
    formatter = logging.Formatter(`%(message)s')
    file_handler.setFormatter(formatter)
    console_handler.setFormatter(formatter)

    # Add handlers to logger
    logger.addHandler(file_handler)
    logger.addHandler(console_handler)

    return logger

# Initialize different LLM clients
def get_llm_client(model_type=``openai''):
    if model_type == ``openai'':
        return OpenAI()
    elif model_type == ``anthropic'':
        return Anthropic(api_key=os.getenv(``ANTHROPIC_API_KEY''))
    elif model_type == ``google'':
        genai.configure(api_key=os.getenv(``GOOGLE_API_KEY''))
        return genai
    elif model_type == ``deepseek'':
        return Deepseek(api_key=os.getenv(``DEEPSEEK_API_KEY''))
    else:
        raise ValueError(f``Unsupported model type: {model_type}'')

# DMU dataset loader
# [You can load TA, ABZ, SWV, YN, and other datasets in similar way]

def load_dmu_dataset(filepath):
    jobs = []
    with open(filepath, `r') as f:
        lines = [line.strip() for line in f if line.strip()]
        n_jobs, n_machines = map(int, lines[0].split())
        for job_idx, line in enumerate(lines[1:]):
            tokens = list(map(int, line.split()))
            steps = [(f``Machine{machine}'', duration) for machine, duration in zip(tokens[::2], tokens[1::2])]
            jobs.append({`name': f`Job{job_idx+1}', `steps': steps})
    return jobs

# After imports, add the list of datasets
DATASETS = [
   ``rcmax_20_15_5'', #Dmu03_
   ``rcmax_20_15_8'', #Dmu04_
   ``rcmax_20_20_7'', #Dmu08_
   ``rcmax_20_20_8'', #Dmu09_
   ``rcmax_30_15_5'', #Dmu13_
   ``rcmax_30_15_4'', #Dmu14_
   ``rcmax_30_20_9'', #Dmu18_
   ``rcmax_30_20_8'', #Dmu19_
   ``rcmax_40_15_10'',#Dmu23_
   ``rcmax_40_15_8'', #Dmu24_
   ``rcmax_40_20_6'', #Dmu28_
   ``rcmax_40_20_2'', #Dmu29_
   ``rcmax_50_15_2'', #Dmu33_
   ``rcmax_50_15_4'', #Dmu34_
   ``rcmax_50_20_6'', #Dmu38_
   ``rcmax_50_20_9'' #Dmu39_
]

# Helper: create a placeholder schedule for demonstration
# In a real system, the agent would compute this based on constraints
step_labels = [`A', `B', `C', `D', `E']
def make_placeholder_schedule(job, offset=0):
    schedule = []
    t = offset
    for idx, (machine, duration) in enumerate(job[`steps']):
        schedule.append({
            `job': job[`name'],
            `step': idx+1,
            `machine': machine,
            `start': t,
            `end': t+duration,
            `precedence': f``After {job[`name']} Step {idx}'' if idx > 0 else None
        })
        t += duration
    return schedule

\end{lstlisting}

\begin{lstlisting}[style=PythonStyle, caption=Collaborative Agents and Prompts (1), label=lst:data_collection]
# Create agents for each job, with run() returning a standardized schedule
class JSSPAgent(Agent):
    def __init__(self, name, backstory, task_description, task_expected_output, model_type=``openai''):
        super().__init__(name, backstory, task_description, task_expected_output)
        self.client = get_llm_client(model_type)
        self.model_type = model_type

    def run(self):
        job_idx = int(self.name.split(`Job')[1][0]) - 1
        job = jobs[job_idx]
        
        # Initialize schedule
        schedule = []
        current_time = 0
        
        # Track machine availability
        machine_availability = {name: 0 for name in machine_names}
        
        for step_idx, (machine, duration) in enumerate(job[`steps']):
            # Find the earliest possible start time considering:
            # 1. Machine availability
            # 2. Previous step completion
            # 3. No overlapping operations on the same machine
            start_time = max(
                machine_availability[machine],
                current_time
            )
            
            # Check for conflicts with other operations
            while True:
                conflict = False
                for existing_schedule in schedule:
                    if (existing_schedule[`machine'] == machine and
                        not (start_time >= existing_schedule[`end'] or 
                             start_time + duration <= existing_schedule[`start'])):
                        conflict = True
                        start_time = existing_schedule[`end']
                        break
                if not conflict:
                    break
            
            # Update machine availability
            machine_availability[machine] = start_time + duration
            
            # Update current time for next step
            current_time = start_time + duration
            
            schedule.append({
                `job': job[`name'],
                `step': step_idx + 1,
                `machine': machine,
                `start': start_time,
                `end': start_time + duration,
                `precedence': f``After {job[`name']} Step {step_idx}'' if step_idx > 0 else None
            })
        
        # Store schedule in agent's context
        self.context = {`schedule': schedule}
        return {`schedule': schedule}

# algorithm 1: tabu search
class SupervisorAgent(Agent):
   ``````
    Supervisor agent that coordinates all job schedules to find the minimum makespan solution.
    Implements multiple scheduling algorithms and selects the best one.
   ''''''
    def __init__(self, name, backstory, task_description, task_expected_output, model_type=``openai''):
        super().__init__(name, backstory, task_description, task_expected_output)
        self.client = get_llm_client(model_type)
        self.model_type = model_type
        self.has_run = False
        self.dataset_name = os.path.splitext(os.path.basename(DMU_FILE))[0]

        
    def schedule_from_order(self, job_order, machine_names):
       ``````Helper function to create a schedule from a job order''''''
        machine_availability = {name: 0 for name in machine_names}
        job_completion = {job[`name']: 0 for job in job_order}
        job_step = {job[`name']: 0 for job in job_order}
        completed_jobs = set()
        schedule = []
        
        while len(completed_jobs) < len(job_order):
            for job in job_order:
                job_name = job[`name']
                if job_name in completed_jobs:
                    continue
                
                current_step = job_step[job_name]
                if current_step >= len(job[`steps']):
                    completed_jobs.add(job_name)
                    continue
                
                machine, duration = job[`steps'][current_step]
                start_time = max(
                    job_completion[job_name],
                    machine_availability[machine]
                )
                
                # Check for conflicts
                while True:
                    conflict = False
                    for existing_op in schedule:
                        if (existing_op[`machine'] == machine and
                            not (start_time >= existing_op[`end'] or 
                                 start_time + duration <= existing_op[`start'])):
                            conflict = True
                            start_time = existing_op[`end']
                            break
                    if not conflict:
                        break
                
                end_time = start_time + duration
                schedule.append({
                    `job': job_name,
                    `step': current_step + 1,
                    `machine': machine,
                    `start': start_time,
                    `end': end_time,
                    `precedence': f``After {job_name} Step {current_step}'' if current_step > 0 else None
                })
                
                machine_availability[machine] = end_time
                job_completion[job_name] = end_time
                job_step[job_name] += 1
                
                if job_step[job_name] >= len(job[`steps']):
                    completed_jobs.add(job_name)
        
        return schedule

    def run(self):
        if self.has_run and hasattr(self, `context') and `schedule' in self.context:
            print(``[SupervisorAgent] Using previously computed schedule'')
            return self.context

        # Create results directory if it doesn't exist
        results_dir = ``./results''
        if not os.path.exists(results_dir):
            os.makedirs(results_dir)
            print(f``Created results directory: {results_dir}'')

        # Create output file
        output_file = f``./results/{self.dataset_name}_dmu.txt''
        with open(output_file, `w') as f:
            f.write(f``=== JSSP Schedule Results for {self.dataset_name} ===\n'')
            f.write(f``Timestamp: {datetime.now().strftime(`%Y-%m-%d %H:%M:%S')}\n\n'')
            print(f``=== JSSP Schedule Results for {self.dataset_name} ==='')
            print(f``Timestamp: {datetime.now().strftime(`%Y-%m-%d %H:%M:%S')}\n'')

        # Get all job schedules from dependencies
        all_schedules = []
        for agent in self.dependencies:
            if hasattr(agent, `context') and isinstance(agent.context, dict) and `schedule' in agent.context:
                all_schedules.extend(agent.context[`schedule'])

        if not all_schedules:
            print(``Warning: No schedules found from job agents'')
            return {`schedule': []}

        # [Run all baselines/algorithms and track their results]
        # [Add your methods here]
        algorithms = {
            `Random': self.schedule_random,
            `LPT': self.schedule_lpt,
            `SPT': self.schedule_spt,
            `STPT': self.schedule_stpt,
            `MPSR': self.schedule_mpsr,
            `DRL-Liu': self.schedule_drl_liu,
            `GP': self.schedule_gp,
            `GEP': self.schedule_gep,
            `Tabu': self.schedule_tabu
        }

        results = {}
        for name, algorithm in algorithms.items():
            print(f``\nRunning {name} algorithm...'')
            schedule = algorithm(jobs, machine_names)
            makespan = max(entry.get(`end', 0) for entry in schedule)
            results[name] = {
                `schedule': schedule,
                `makespan': makespan
            }
            print(f``{name} makespan: {makespan}'')

        # Find the best algorithm
        best_algorithm = min(results.items(), key=lambda x: x[1][`makespan'])
        best_name, best_result = best_algorithm
        
        print(f``\nBest algorithm: {best_name} with makespan {best_result[`makespan']}'')
        
        # Calculate upper bound
        job_sums = {}
        machine_sums = {name: 0 for name in machine_names}
        for job in jobs:
            job_sum = sum(duration for _, duration in job[`steps'])
            job_sums[job[`name']] = job_sum
        for entry in best_result[`schedule']:
            machine = entry.get(`machine')
            duration = entry.get(`end', 0) - entry.get(`start', 0)
            if machine in machine_sums:
                machine_sums[machine] += duration
        ub = max(max(job_sums.values()), max(machine_sums.values()))
        
        # Write results to file and print to console
        results_text = [
           ``\n=== Final Results ===``,
            f``Best Algorithm: {best_name}'',
            f``Minimum Makespan found: {best_result[`makespan']}'',
            f``Upper Bound (UB): {ub}'',
            f``Gap to UB: {best_result[`makespan'] - ub}\n'',
           ``=== Algorithm Comparison ===``
        ]
        
        for name, result in results.items():
            results_text.append(f``{name}: {result[`makespan']}'')
        
        results_text.append(``\n=== Detailed Schedule ==='')

        with open(output_file, `a') as f:
            for line in results_text:
                print(line)
                f.write(line + ''\n'')
            
            # Write detailed schedule
            for entry in best_result[`schedule']:
                schedule_line = f``Job: {entry[`job']}, Step: {entry[`step']}, Machine: {entry[`machine']}, Start: {entry[`start']}, End: {entry[`end']}''
                print(schedule_line)
                f.write(schedule_line + ''\n'')

        self.has_run = True
        self.context = {`schedule': best_result[`schedule']}
        return {`schedule': best_result[`schedule']}
\end{lstlisting}

\begin{lstlisting}[style=PythonStyle, caption=Collaborative Agents and Prompts (2), label=lst:data_collection]
# Modify the main execution section at the bottom of the file
if __name__ == ``__main__'':
    # Process each dataset
    for dataset in DATASETS:
        # Setup logging for this dataset
        logger = setup_logging(dataset)
        
        logger.info(f``\n{`='*50}'')
        logger.info(f``Processing dataset: {dataset}'')
        logger.info(f``{`='*50}\n'')

        # Update DMU file path for current dataset
        DMU_FILE = os.path.join(project_root, `applications', `DMU', f`{dataset}.txt')
        
        # Load jobs for current dataset
        jobs = load_dmu_dataset(DMU_FILE)
        
        # After loading jobs
        all_machine_indices = set()
        for job in jobs:
            for machine, _ in job[`steps']:
                idx = int(machine.replace(`Machine', `'))
                all_machine_indices.add(idx)
        machine_names = [f``Machine{idx}'' for idx in sorted(all_machine_indices)]

        # Create agents for each job
        agents = []
        for job in jobs:
            agent = JSSPAgent(
                name=f``{job[`name']} Agent'',
                backstory=f``Agent for {job[`name']} scheduling.'',
                task_description=f``Schedule steps for {job[`name']} on required machines with precedence.'',
                task_expected_output=f``Step schedule for {job[`name']} respecting machine and precedence constraints.'',
                model_type=``openai''
            )
            agents.append(agent)
        
        # Add validation agent
        # validation_agent = JSSPValidationAgent(
        #     name=``JSSP Validation Agent'',
        #     backstory=``Validates JSSP schedules for constraint violations.'',
        #     task_description=``Check all schedules for machine constraints, precedence constraints, and makespan validity.'',
        #     task_expected_output=``Validation results with any detected violations.'',
        #     model_type=``openai''
        # )

        # Add supervisor agent
        supervisor_agent = SupervisorAgent(
            name=``Supervisor Agent'',
            backstory=``Supervisor agent that coordinates all job schedules to find the minimum makespan solution.'',
            task_description=``````Find the minimum makespan schedule for all jobs while strictly following these rules:
1. Each job steps must be completed in strict order (e.g., Job1 step 2 can only start after step 1 is completed).
2. Each machine can only process one job step at a time (e.g., if MachineA is processing Job1 step 1 from time 0-3, it cannot process any other job steps during that time).

The goal is to minimize the total completion time (makespan) while ensuring all jobs are completed and all constraints are satisfied.'''''',
            task_expected_output=``A complete schedule with minimum makespan that satisfies all constraints.'',
            model_type=``openai''
        )

        agents.extend([supervisor_agent])

        # Only job agents as initial nodes
        nodes = [{`agent': agent, `dependencies': []} for agent in agents if isinstance(agent, JSSPAgent)]

        # Supervisor depends on all job agents
        nodes.append({`agent': supervisor_agent, `dependencies': [agent.name for agent in agents if isinstance(agent, JSSPAgent)]})

        # Validation agent depends on supervisor
        # nodes.append({`agent': validation_agent, `dependencies': [supervisor_agent.name]})

        task_spec = {
            `nodes': nodes,
            `edges': [],
            `jobs': jobs,
            `disruptions': [],
            `rules': [
                `Each job must perform its steps strictly in order.',
                `Each machine can only handle one operation at a time.',
                `No two operations use the same machine at the same time.'
            ]
        }

        # Initialize MAPLE
        maple = MAPLE(task_spec)

        # Run MAPLE
        maple.run(with_rollback=True, validate=True)

        # Extract and print overall schedule from supervisor agent
        context = maple.executor.context
        supervisor_output = context.get(supervisor_agent.name, {})

        logger.info(f``=== Results for {dataset} ==='')
        logger.info(f``Timestamp: {datetime.now().strftime(`%Y-%m-%d %H:%M:%S')}\n'')

        if isinstance(supervisor_output, dict) and `schedule' in supervisor_output:
            all_schedules = supervisor_output[`schedule']
            # Sort by start time for overall schedule
            all_schedules.sort(key=lambda x: (x.get(`start', 0), x.get(`machine', `'), x.get(`job', `')))
            
            # Write detailed schedule
            logger.info(``| Job  | Step | Machine  | Time Slot | Precedence Constraints      |'')
            logger.info(``|------|------|----------|-----------|----------------------------|'')
            for entry in all_schedules:
                job = entry.get(`job', `?')
                step = entry.get(`step', `?')
                machine = entry.get(`machine', `?')
                start = entry.get(`start', `?')
                end = entry.get(`end', `?')
                prec = entry.get(`precedence', `None')
                logger.info(f``| {job} | {step} | {machine} | {start}-{end} | {prec} |'')
            
            # Calculate and write makespan
            makespan = max(entry.get(`end', 0) for entry in all_schedules)
            logger.info(f``\nBest static makespan for {dataset}: {makespan}'')

            # Write Gantt chart
            logger.info(``\n=== JSSP Gantt Chart/Table (Textual) ==='')
            logger.info(``Time | Machine | Job | Step'')
            for entry in all_schedules:
                logger.info(f``{entry.get(`start', `?')}-{entry.get(`end', `?')} | {entry.get(`machine', `?')} | {entry.get(`job', `?')} | {entry.get(`step', `?')}'')
        else:
            logger.info(f``\nNo detailed schedules found for {dataset}'')

        logger.info(f``\nCompleted processing {dataset}'')
        logger.info(f``{`='*50}\n'')

        # Remove handlers to avoid duplicate logging in next iteration
        for handler in logger.handlers[:]:
            handler.close()
            logger.removeHandler(handler) 
            
\end{lstlisting}

\subsection{Meta Plan Execution without Disruption (P11)}

This section depicts the execution of the meta-plan (workflow) of JSSP problem for data sample rcmax\_20\_15\_5.

\begin{lstlisting}[style=JSONStyle, caption=Sample JSON Output, label=lst:data_collection]
=== Results for rcmax_20_15_5 ===
Timestamp: 2025-05-09 12:21:57

| Job  | Step | Machine  | Time Slot | Precedence Constraints      |
|------|------|----------|-----------|----------------------------|
| Job5 | 1 | Machine1 | 0-199 | None |
| Job1 | 1 | Machine11 | 0-84 | None |
| Job20 | 1 | Machine12 | 0-185 | None |
| Job7 | 1 | Machine13 | 0-194 | None |
| Job2 | 1 | Machine14 | 0-95 | None |
| Job18 | 1 | Machine4 | 0-42 | None |
| Job6 | 1 | Machine6 | 0-113 | None |
| Job9 | 1 | Machine7 | 0-124 | None |
| Job17 | 1 | Machine8 | 0-104 | None |
| Job11 | 1 | Machine4 | 42-132 | None |
| Job2 | 2 | Machine0 | 95-186 | After Job2 Step 1 |
| Job14 | 1 | Machine14 | 95-104 | None |
| Job10 | 1 | Machine14 | 104-272 | None |
| Job14 | 2 | Machine2 | 104-267 | After Job14 Step 1 |
| Job6 | 2 | Machine3 | 113-222 | After Job6 Step 1 |
| Job8 | 1 | Machine6 | 113-216 | None |
| Job16 | 1 | Machine7 | 124-169 | None |
| Job12 | 1 | Machine4 | 132-177 | None |
| Job16 | 2 | Machine5 | 169-339 | After Job16 Step 1 |
| Job13 | 1 | Machine7 | 169-276 | None |
| Job1 | 2 | Machine12 | 185-304 | After Job1 Step 1 |
| Job20 | 2 | Machine8 | 185-204 | After Job20 Step 1 |
| Job12 | 2 | Machine0 | 186-231 | After Job12 Step 1 |
| Job7 | 2 | Machine11 | 194-327 | After Job7 Step 1 |
| Job4 | 1 | Machine13 | 194-313 | None |
| Job2 | 3 | Machine8 | 204-357 | After Job2 Step 2 |
| Job8 | 2 | Machine4 | 216-309 | After Job8 Step 1 |
| Job11 | 2 | Machine3 | 222-421 | After Job11 Step 1 |
| Job18 | 2 | Machine14 | 272-334 | After Job18 Step 1 |
| Job13 | 2 | Machine6 | 276-413 | After Job13 Step 1 |
| Job15 | 1 | Machine7 | 276-463 | None |
| Job1 | 3 | Machine2 | 304-432 | After Job1 Step 2 |
| Job3 | 1 | Machine13 | 313-404 | None |
| Job5 | 2 | Machine11 | 327-340 | After Job5 Step 1 |
| Job7 | 3 | Machine12 | 327-444 | After Job7 Step 2 |
| Job4 | 2 | Machine5 | 339-527 | After Job4 Step 1 |
| Job5 | 3 | Machine10 | 340-403 | After Job5 Step 2 |
| Job9 | 2 | Machine11 | 340-525 | After Job9 Step 1 |
| Job14 | 3 | Machine10 | 403-507 | After Job14 Step 2 |
| Job5 | 4 | Machine8 | 403-461 | After Job5 Step 3 |
| Job18 | 3 | Machine13 | 404-490 | After Job18 Step 2 |
| Job3 | 2 | Machine3 | 421-483 | After Job3 Step 1 |
| Job13 | 3 | Machine2 | 432-446 | After Job13 Step 2 |
| Job11 | 3 | Machine12 | 444-629 | After Job11 Step 2 |
| Job13 | 4 | Machine9 | 446-559 | After Job13 Step 3 |
| Job15 | 2 | Machine4 | 463-518 | After Job15 Step 1 |
| Job19 | 1 | Machine7 | 463-623 | None |
| Job20 | 3 | Machine3 | 483-494 | After Job20 Step 2 |
| Job16 | 3 | Machine3 | 494-629 | After Job16 Step 2 |
| Job14 | 4 | Machine13 | 507-527 | After Job14 Step 3 |
| Job9 | 3 | Machine6 | 525-678 | After Job9 Step 2 |
| Job6 | 3 | Machine5 | 527-712 | After Job6 Step 2 |
| Job19 | 2 | Machine1 | 623-742 | After Job19 Step 1 |
| Job10 | 2 | Machine7 | 623-628 | After Job10 Step 1 |
| Job10 | 3 | Machine4 | 628-645 | After Job10 Step 2 |
| Job17 | 2 | Machine7 | 628-703 | After Job17 Step 1 |
| Job8 | 3 | Machine12 | 629-650 | After Job8 Step 2 |
| Job4 | 3 | Machine3 | 629-672 | After Job4 Step 2 |
| Job3 | 3 | Machine4 | 645-818 | After Job3 Step 2 |
| Job10 | 4 | Machine9 | 645-831 | After Job10 Step 3 |
| Job8 | 4 | Machine10 | 650-798 | After Job8 Step 3 |
| Job1 | 4 | Machine3 | 672-816 | After Job1 Step 3 |
| Job2 | 4 | Machine6 | 678-787 | After Job2 Step 3 |
| Job17 | 3 | Machine0 | 703-886 | After Job17 Step 2 |
| Job12 | 3 | Machine7 | 703-860 | After Job12 Step 2 |
| Job15 | 3 | Machine5 | 712-788 | After Job15 Step 2 |
| Job6 | 4 | Machine8 | 712-771 | After Job6 Step 3 |
| Job19 | 3 | Machine14 | 742-835 | After Job19 Step 2 |
| Job6 | 5 | Machine1 | 771-774 | After Job6 Step 4 |
| Job11 | 4 | Machine8 | 771-865 | After Job11 Step 3 |
| Job16 | 4 | Machine6 | 787-859 | After Job16 Step 3 |
| Job15 | 4 | Machine2 | 788-844 | After Job15 Step 3 |
| Job9 | 4 | Machine5 | 788-931 | After Job9 Step 3 |
| Job3 | 4 | Machine11 | 818-885 | After Job3 Step 3 |
| Job5 | 5 | Machine4 | 818-873 | After Job5 Step 4 |
| Job7 | 4 | Machine14 | 835-848 | After Job7 Step 3 |
| Job19 | 4 | Machine3 | 835-857 | After Job19 Step 3 |
| Job2 | 5 | Machine2 | 844-1026 | After Job2 Step 4 |
| Job12 | 4 | Machine14 | 860-873 | After Job12 Step 3 |
| Job20 | 4 | Machine7 | 860-947 | After Job20 Step 3 |
| Job11 | 5 | Machine6 | 865-905 | After Job11 Step 4 |
| Job4 | 4 | Machine8 | 865-883 | After Job4 Step 3 |
| Job12 | 5 | Machine1 | 873-999 | After Job12 Step 4 |
| Job5 | 6 | Machine9 | 873-955 | After Job5 Step 5 |
| Job1 | 5 | Machine8 | 883-1060 | After Job1 Step 4 |
| Job18 | 4 | Machine0 | 886-1004 | After Job18 Step 3 |
| Job17 | 4 | Machine14 | 886-1038 | After Job17 Step 3 |
| Job10 | 5 | Machine6 | 905-1038 | After Job10 Step 4 |
| Job8 | 5 | Machine5 | 931-997 | After Job8 Step 4 |
| Job20 | 5 | Machine11 | 947-1118 | After Job20 Step 4 |
| Job7 | 5 | Machine7 | 947-1058 | After Job7 Step 4 |
| Job6 | 6 | Machine9 | 955-979 | After Job6 Step 5 |
| Job8 | 6 | Machine3 | 997-1026 | After Job8 Step 5 |
| Job5 | 7 | Machine5 | 997-1019 | After Job5 Step 6 |
| Job13 | 5 | Machine1 | 999-1137 | After Job13 Step 4 |
| Job9 | 5 | Machine0 | 1004-1034 | After Job9 Step 4 |
| Job18 | 5 | Machine12 | 1004-1132 | After Job18 Step 4 |
| Job2 | 6 | Machine10 | 1026-1073 | After Job2 Step 5 |
| Job14 | 5 | Machine0 | 1034-1055 | After Job14 Step 4 |
| Job17 | 5 | Machine2 | 1038-1204 | After Job17 Step 4 |
| Job3 | 5 | Machine0 | 1055-1191 | After Job3 Step 4 |
| Job14 | 6 | Machine3 | 1055-1103 | After Job14 Step 5 |
| Job19 | 5 | Machine8 | 1060-1228 | After Job19 Step 4 |
| Job2 | 7 | Machine7 | 1073-1171 | After Job2 Step 6 |
| Job16 | 5 | Machine11 | 1118-1174 | After Job16 Step 4 |
| Job20 | 6 | Machine4 | 1118-1132 | After Job20 Step 5 |
| Job4 | 5 | Machine12 | 1132-1155 | After Job4 Step 4 |
| Job20 | 7 | Machine13 | 1132-1197 | After Job20 Step 6 |
| Job15 | 5 | Machine1 | 1137-1196 | After Job15 Step 4 |
| Job4 | 6 | Machine14 | 1155-1213 | After Job4 Step 5 |
| Job12 | 6 | Machine11 | 1174-1218 | After Job12 Step 5 |
| Job1 | 6 | Machine0 | 1191-1342 | After Job1 Step 5 |
| Job3 | 6 | Machine10 | 1191-1331 | After Job3 Step 5 |
| Job8 | 7 | Machine1 | 1196-1207 | After Job8 Step 6 |
| Job15 | 6 | Machine12 | 1196-1207 | After Job15 Step 5 |
| Job18 | 6 | Machine2 | 1204-1357 | After Job18 Step 5 |
| Job14 | 7 | Machine1 | 1207-1338 | After Job14 Step 6 |
| Job15 | 7 | Machine13 | 1207-1281 | After Job15 Step 6 |
| Job12 | 7 | Machine5 | 1218-1370 | After Job12 Step 6 |
| Job19 | 6 | Machine12 | 1228-1366 | After Job19 Step 5 |
| Job9 | 6 | Machine8 | 1228-1255 | After Job9 Step 5 |
| Job13 | 6 | Machine8 | 1255-1437 | After Job13 Step 5 |
| Job17 | 6 | Machine10 | 1331-1341 | After Job17 Step 5 |
| Job20 | 8 | Machine1 | 1338-1399 | After Job20 Step 7 |
| Job17 | 7 | Machine11 | 1341-1463 | After Job17 Step 6 |
| Job11 | 6 | Machine0 | 1342-1434 | After Job11 Step 5 |
| Job1 | 7 | Machine9 | 1342-1480 | After Job1 Step 6 |
| Job7 | 6 | Machine2 | 1357-1483 | After Job7 Step 5 |
| Job3 | 7 | Machine12 | 1366-1481 | After Job3 Step 6 |
| Job12 | 8 | Machine10 | 1370-1518 | After Job12 Step 7 |
| Job16 | 6 | Machine0 | 1434-1580 | After Job16 Step 5 |
| Job13 | 7 | Machine4 | 1437-1616 | After Job13 Step 6 |
| Job18 | 7 | Machine8 | 1437-1571 | After Job18 Step 6 |
| Job2 | 8 | Machine11 | 1463-1517 | After Job2 Step 7 |
| Job1 | 8 | Machine6 | 1480-1496 | After Job1 Step 7 |
| Job8 | 8 | Machine9 | 1480-1484 | After Job8 Step 7 |
| Job5 | 8 | Machine12 | 1481-1664 | After Job5 Step 7 |
| Job11 | 7 | Machine2 | 1483-1629 | After Job11 Step 6 |
| Job18 | 8 | Machine6 | 1571-1727 | After Job18 Step 7 |
| Job7 | 7 | Machine8 | 1571-1672 | After Job7 Step 6 |
| Job10 | 6 | Machine0 | 1580-1615 | After Job10 Step 5 |
| Job6 | 7 | Machine0 | 1615-1686 | After Job6 Step 6 |
| Job10 | 7 | Machine13 | 1615-1716 | After Job10 Step 6 |
| Job13 | 8 | Machine11 | 1616-1723 | After Job13 Step 7 |
| Job2 | 9 | Machine4 | 1616-1775 | After Job2 Step 8 |
| Job9 | 7 | Machine2 | 1629-1698 | After Job9 Step 6 |
| Job11 | 8 | Machine7 | 1629-1719 | After Job11 Step 7 |
| Job5 | 9 | Machine3 | 1664-1707 | After Job5 Step 8 |
| Job16 | 7 | Machine8 | 1672-1862 | After Job16 Step 6 |
| Job6 | 8 | Machine14 | 1686-1784 | After Job6 Step 7 |
| Job4 | 7 | Machine2 | 1698-1834 | After Job4 Step 6 |
| Job9 | 8 | Machine9 | 1698-1828 | After Job9 Step 7 |
| Job19 | 7 | Machine13 | 1716-1878 | After Job19 Step 6 |
| Job14 | 8 | Machine11 | 1723-1732 | After Job14 Step 7 |
| Job7 | 8 | Machine6 | 1727-1765 | After Job7 Step 7 |
| Job18 | 9 | Machine11 | 1732-1812 | After Job18 Step 8 |
| Job14 | 9 | Machine6 | 1765-1890 | After Job14 Step 8 |
| Job1 | 9 | Machine14 | 1784-1979 | After Job1 Step 8 |
| Job9 | 9 | Machine12 | 1828-1881 | After Job9 Step 8 |
| Job11 | 9 | Machine9 | 1828-1959 | After Job11 Step 8 |
| Job4 | 8 | Machine0 | 1834-1888 | After Job4 Step 7 |
| Job16 | 8 | Machine2 | 1862-1919 | After Job16 Step 7 |
| Job15 | 8 | Machine8 | 1862-1864 | After Job15 Step 7 |
| Job15 | 9 | Machine3 | 1864-2058 | After Job15 Step 8 |
| Job10 | 8 | Machine8 | 1864-2036 | After Job10 Step 7 |
| Job17 | 8 | Machine13 | 1878-1910 | After Job17 Step 7 |
| Job19 | 8 | Machine5 | 1878-1943 | After Job19 Step 7 |
| Job9 | 10 | Machine4 | 1881-2070 | After Job9 Step 9 |
| Job7 | 9 | Machine0 | 1888-2072 | After Job7 Step 8 |
| Job4 | 9 | Machine6 | 1890-2084 | After Job4 Step 8 |
| Job6 | 9 | Machine13 | 1910-1942 | After Job6 Step 8 |
| Job3 | 8 | Machine2 | 1919-2102 | After Job3 Step 7 |
| Job6 | 10 | Machine10 | 1942-2044 | After Job6 Step 9 |
| Job8 | 9 | Machine13 | 1942-1970 | After Job8 Step 8 |
| Job2 | 10 | Machine9 | 1959-2082 | After Job2 Step 9 |
| Job16 | 9 | Machine13 | 1970-2118 | After Job16 Step 8 |
| Job1 | 10 | Machine5 | 1979-2072 | After Job1 Step 9 |
| Job6 | 11 | Machine12 | 2044-2063 | After Job6 Step 10 |
| Job10 | 9 | Machine3 | 2058-2114 | After Job10 Step 8 |
| Job20 | 9 | Machine0 | 2072-2106 | After Job20 Step 8 |
| Job18 | 10 | Machine5 | 2072-2177 | After Job18 Step 9 |
| Job7 | 10 | Machine9 | 2082-2217 | After Job7 Step 9 |
| Job4 | 10 | Machine1 | 2084-2119 | After Job4 Step 9 |
| Job19 | 9 | Machine6 | 2084-2140 | After Job19 Step 8 |
| Job3 | 9 | Machine14 | 2102-2288 | After Job3 Step 8 |
| Job12 | 9 | Machine2 | 2102-2224 | After Job12 Step 8 |
| Job5 | 10 | Machine0 | 2106-2263 | After Job5 Step 9 |
| Job10 | 10 | Machine10 | 2114-2240 | After Job10 Step 9 |
| Job13 | 9 | Machine13 | 2118-2236 | After Job13 Step 8 |
| Job17 | 9 | Machine6 | 2140-2234 | After Job17 Step 8 |
| Job11 | 10 | Machine5 | 2177-2234 | After Job11 Step 9 |
| Job18 | 11 | Machine7 | 2177-2193 | After Job18 Step 10 |
| Job18 | 12 | Machine3 | 2193-2379 | After Job18 Step 11 |
| Job7 | 11 | Machine1 | 2217-2316 | After Job7 Step 10 |
| Job20 | 10 | Machine9 | 2217-2310 | After Job20 Step 9 |
| Job8 | 10 | Machine2 | 2224-2317 | After Job8 Step 9 |
| Job12 | 10 | Machine6 | 2234-2392 | After Job12 Step 9 |
| Job1 | 11 | Machine13 | 2236-2343 | After Job1 Step 10 |
| Job13 | 10 | Machine5 | 2236-2408 | After Job13 Step 9 |
| Job16 | 10 | Machine14 | 2288-2327 | After Job16 Step 9 |
| Job20 | 11 | Machine10 | 2310-2464 | After Job20 Step 10 |
| Job14 | 10 | Machine9 | 2310-2411 | After Job14 Step 9 |
| Job8 | 11 | Machine11 | 2317-2509 | After Job8 Step 10 |
| Job10 | 11 | Machine2 | 2317-2392 | After Job10 Step 10 |
| Job16 | 11 | Machine12 | 2327-2490 | After Job16 Step 10 |
| Job5 | 11 | Machine14 | 2327-2352 | After Job5 Step 10 |
| Job2 | 11 | Machine13 | 2343-2348 | After Job2 Step 10 |
| Job11 | 11 | Machine13 | 2348-2483 | After Job11 Step 10 |
| Job9 | 11 | Machine14 | 2352-2438 | After Job9 Step 10 |
| Job12 | 11 | Machine8 | 2392-2540 | After Job12 Step 10 |
| Job3 | 10 | Machine5 | 2408-2414 | After Job3 Step 9 |
| Job14 | 11 | Machine4 | 2411-2517 | After Job14 Step 10 |
| Job15 | 10 | Machine9 | 2411-2424 | After Job15 Step 9 |
| Job3 | 11 | Machine1 | 2414-2604 | After Job3 Step 10 |
| Job19 | 10 | Machine9 | 2424-2595 | After Job19 Step 9 |
| Job13 | 11 | Machine14 | 2438-2595 | After Job13 Step 10 |
| Job15 | 11 | Machine10 | 2464-2568 | After Job15 Step 10 |
| Job20 | 12 | Machine6 | 2464-2531 | After Job20 Step 11 |
| Job5 | 12 | Machine13 | 2483-2543 | After Job5 Step 11 |
| Job8 | 12 | Machine0 | 2509-2576 | After Job8 Step 11 |
| Job4 | 11 | Machine4 | 2517-2557 | After Job4 Step 10 |
| Job14 | 12 | Machine8 | 2540-2735 | After Job14 Step 11 |
| Job12 | 12 | Machine13 | 2543-2646 | After Job12 Step 11 |
| Job4 | 12 | Machine7 | 2557-2589 | After Job4 Step 11 |
| Job7 | 12 | Machine10 | 2568-2660 | After Job7 Step 11 |
| Job15 | 12 | Machine6 | 2568-2715 | After Job15 Step 11 |
| Job19 | 11 | Machine11 | 2595-2615 | After Job19 Step 10 |
| Job17 | 10 | Machine9 | 2595-2756 | After Job17 Step 9 |
| Job1 | 12 | Machine1 | 2604-2626 | After Job1 Step 11 |
| Job6 | 12 | Machine11 | 2615-2635 | After Job6 Step 11 |
| Job19 | 12 | Machine2 | 2615-2623 | After Job19 Step 11 |
| Job20 | 13 | Machine2 | 2623-2637 | After Job20 Step 12 |
| Job16 | 12 | Machine1 | 2626-2640 | After Job16 Step 11 |
| Job11 | 12 | Machine11 | 2635-2825 | After Job11 Step 11 |
| Job6 | 13 | Machine4 | 2635-2747 | After Job6 Step 12 |
| Job10 | 12 | Machine1 | 2640-2733 | After Job10 Step 11 |
| Job9 | 12 | Machine10 | 2660-2738 | After Job9 Step 11 |
| Job3 | 12 | Machine6 | 2715-2888 | After Job3 Step 11 |
| Job18 | 13 | Machine1 | 2733-2817 | After Job18 Step 12 |
| Job14 | 13 | Machine7 | 2735-2896 | After Job14 Step 12 |
| Job8 | 13 | Machine8 | 2735-2831 | After Job8 Step 12 |
| Job1 | 13 | Machine10 | 2738-2875 | After Job1 Step 12 |
| Job16 | 13 | Machine4 | 2747-2915 | After Job16 Step 12 |
| Job17 | 11 | Machine12 | 2756-2906 | After Job17 Step 10 |
| Job4 | 13 | Machine9 | 2756-2940 | After Job4 Step 12 |
| Job11 | 13 | Machine1 | 2825-3017 | After Job11 Step 12 |
| Job15 | 13 | Machine11 | 2825-2991 | After Job15 Step 12 |
| Job18 | 14 | Machine10 | 2875-2917 | After Job18 Step 13 |
| Job5 | 13 | Machine6 | 2888-3038 | After Job5 Step 12 |
| Job2 | 12 | Machine12 | 2906-2911 | After Job2 Step 11 |
| Job17 | 12 | Machine3 | 2906-2907 | After Job17 Step 11 |
| Job12 | 13 | Machine3 | 2907-2976 | After Job12 Step 12 |
| Job13 | 12 | Machine12 | 2911-3089 | After Job13 Step 11 |
| Job2 | 13 | Machine5 | 2911-3052 | After Job2 Step 12 |
| Job19 | 13 | Machine4 | 2915-3052 | After Job19 Step 12 |
| Job3 | 13 | Machine9 | 2940-3079 | After Job3 Step 12 |
| Job4 | 14 | Machine11 | 2991-3103 | After Job4 Step 13 |
| Job9 | 13 | Machine1 | 3017-3172 | After Job9 Step 12 |
| Job11 | 14 | Machine10 | 3017-3073 | After Job11 Step 13 |
| Job5 | 14 | Machine7 | 3038-3050 | After Job5 Step 13 |
| Job5 | 15 | Machine2 | 3050-3165 | After Job5 Step 14 |
| Job6 | 14 | Machine7 | 3050-3064 | After Job6 Step 13 |
| Job1 | 14 | Machine4 | 3052-3148 | After Job1 Step 13 |
| Job7 | 13 | Machine5 | 3052-3198 | After Job7 Step 12 |
| Job8 | 14 | Machine7 | 3064-3080 | After Job8 Step 13 |
| Job11 | 15 | Machine14 | 3073-3176 | After Job11 Step 14 |
| Job3 | 14 | Machine8 | 3079-3107 | After Job3 Step 13 |
| Job16 | 14 | Machine9 | 3079-3180 | After Job16 Step 13 |
| Job10 | 13 | Machine12 | 3089-3156 | After Job10 Step 12 |
| Job13 | 13 | Machine3 | 3089-3216 | After Job13 Step 12 |
| Job1 | 15 | Machine7 | 3148-3169 | After Job1 Step 14 |
| Job10 | 14 | Machine11 | 3156-3265 | After Job10 Step 13 |
| Job12 | 14 | Machine12 | 3156-3249 | After Job12 Step 13 |
| Job6 | 15 | Machine2 | 3165-3204 | After Job6 Step 14 |
| Job3 | 15 | Machine7 | 3169-3352 | After Job3 Step 14 |
| Job17 | 13 | Machine1 | 3172-3270 | After Job17 Step 12 |
| Job9 | 14 | Machine13 | 3172-3259 | After Job9 Step 13 |
| Job8 | 15 | Machine14 | 3176-3240 | After Job8 Step 14 |
| Job16 | 15 | Machine10 | 3180-3279 | After Job16 Step 14 |
| Job18 | 15 | Machine9 | 3180-3309 | After Job18 Step 14 |
| Job20 | 14 | Machine5 | 3198-3334 | After Job20 Step 13 |
| Job13 | 14 | Machine0 | 3216-3250 | After Job13 Step 13 |
| Job7 | 14 | Machine3 | 3216-3260 | After Job7 Step 13 |
| Job15 | 14 | Machine0 | 3250-3284 | After Job15 Step 13 |
| Job2 | 14 | Machine1 | 3270-3349 | After Job2 Step 13 |
| Job17 | 14 | Machine4 | 3270-3383 | After Job17 Step 13 |
| Job13 | 15 | Machine10 | 3279-3361 | After Job13 Step 14 |
| Job19 | 14 | Machine0 | 3284-3477 | After Job19 Step 13 |
| Job12 | 15 | Machine9 | 3309-3501 | After Job12 Step 14 |
| Job20 | 15 | Machine14 | 3334-3361 | After Job20 Step 14 |
| Job14 | 14 | Machine5 | 3334-3408 | After Job14 Step 13 |
| Job2 | 15 | Machine3 | 3349-3509 | After Job2 Step 14 |
| Job4 | 15 | Machine10 | 3361-3547 | After Job4 Step 14 |
| Job15 | 15 | Machine14 | 3361-3479 | After Job15 Step 14 |
| Job7 | 15 | Machine4 | 3383-3541 | After Job7 Step 14 |
| Job14 | 15 | Machine12 | 3408-3523 | After Job14 Step 14 |
| Job10 | 15 | Machine5 | 3408-3535 | After Job10 Step 14 |
| Job9 | 15 | Machine3 | 3509-3623 | After Job9 Step 14 |
| Job17 | 15 | Machine5 | 3535-3561 | After Job17 Step 14 |
| Job19 | 15 | Machine10 | 3547-3639 | After Job19 Step 14 |

Best static makespan for rcmax_20_15_5: 3639
\end{lstlisting}

\subsection{ALAS (ours) Implementation (P11)}
Please refer to our paper~\cite{chang2025ALAS} for implementation of the ALAS/MAPLE framework.

\section*{Urban Ride Sharing, Problems P3 and P4}

For a benchmark suite, we encourage users to
devise novel methodologies to solve the problems. However, for this illustrative example of Urban Ride Sharing (URS) (P3 and P4), we employ a customized specification language that extends PDDL (Planning Domain Definition Language) and workflow networks to support:

\begin{itemize}
    \item In sequential planing (P3), we provide static constraints and real-time updates.
    \item In reactive planning (P4), we provide dynamic constraints and real-time updates.
    \item Integration with streaming data sources.
    \item Explicit representation of uncertainty.
    \item Temporal and spatial dependencies.
\end{itemize}

The solver, whether manual or automatic, approaches the planning problem in three steps:
\begin{enumerate}
    \item Convert the problem statement into a formal specification:
    \begin{itemize}
        \item Key objectives and constraints.
        \item Required resources and their capabilities.
        \item Performance metrics and success criteria.
        \item Temporal and spatial dependencies.
    \end{itemize}
    
    \item Transform the specification into a workflow graph:
    \begin{itemize}
        \item Nodes represent processing stages, decision points, or actions.
        \item Edges capture dependencies, data flow, and execution sequence.
        \item Agents are assigned to both nodes and edges.
        \item Concrete parameters and thresholds are specified.
    \end{itemize}
    
    \item Select and apply solving algorithms:
    \begin{itemize}
        \item Test multiple solution approaches (e.g., dynamic programming, Monte Carlo).
        \item Evaluate solutions against specified metrics.
        \item Select and validate the best-performing solution.
        \item Present results with performance analyses.
    \end{itemize}
\end{enumerate}

\subsection{URS Illustration Example}
We walk through these three steps to solve the URS problem.

\subsubsection{URS Problem Specification}Already presented in Table~\ref{tab:appURS}.

\subsubsection{URS Workflow}

Figure~\ref{fig:URS-DirectedGraph} shows
the workflow representation of the URS problem.

\begin{figure}[th!]
\vspace{-.1in}
    \centering
    \begin{tikzpicture}[
        location/.style={
            circle,
            draw=black,
            fill=white!20,
            minimum size=0.8cm,
            text centered,
            font=\large
        },
        airport/.style={
            rectangle,
            rounded corners,
            draw=black,
            fill=cyan!20,
            minimum width=2.0cm,
            minimum height=1.0cm,
            text centered,
            font=\large
        },
        passenger/.style={
            font=\normalsize\color{red}
        },
        vehicle/.style={
            font=\normalsize\color{blue}
        },
        urban_path/.style={thick},
        airport_path/.style={thick}
    ]
        
    \node[location, fill=white, draw, inner sep=2pt, minimum size=28pt] (A) at (0,1.05) {A};
    \node[location, fill=white, draw, inner sep=2pt, minimum size=28pt] (B) at (2.0,2) {B};
    \node[location, fill=white, draw, inner sep=2pt, minimum size=28pt] (C) at (2.0,0.2) {C};
    \node[location, fill=white, draw, inner sep=2pt, minimum size=28pt] (D) at (0,-0.4) {D};
    \node[location, fill=white, draw, inner sep=2pt, minimum size=28pt] (E) at (-2.0,0.2) {E};
    \node[location, fill=white, draw, inner sep=2pt, minimum size=28pt] (F) at (-2.0,2) {F};

        \node[airport] (G) at (3.8,1) {G (BOS)};
        
        \foreach \from/\to in {A/B, B/C, C/D, D/E, E/F, F/A, A/C, B/D, C/E, D/F, E/A, F/B}
            \draw[urban_path] (\from) -- (\to) 
            node[midway,fill=none] {};
            
        \draw[airport_path] (A) -- (G) node[pos=0.72,sloped,above,fill=none] {};
        \draw[airport_path] (B) -- (G) node[pos=0.50,sloped,above,fill=none] {};
        \draw[airport_path] (C) -- (G) node[pos=0.50,sloped,above,fill=none] {};
        \draw[airport_path] (E) -- (G) node[pos=0.78,sloped,above,fill=none] {};
        \draw[airport_path] (F) -- (G) node[pos=0.78,sloped,above,fill=none] {};
        
        \node[passenger] at ($(A)+(0.5,0.6)$) {$r_1$ (8:45)};
        \node[passenger] at ($(B)+(0.5,0.6)$) {$r_2$ (8:50)};
        \node[passenger] at ($(C)+(0.5,-0.6)$) {$r_3$ (8:55)};
        \node[passenger] at ($(D)+(0.5,-0.6)$) {$r_4$ (9:00)};
        
        \node[vehicle] at ($(A)+(-0.5,0.6)$) {$k_1$ [2]};
        \node[vehicle] at ($(C)+(-0.5,-0.6)$) {$k_2$ [2]};
        \node[vehicle] at ($(E)+(-0.5,-0.6)$) {$k_3$ [2]};
        
    \end{tikzpicture}
    \vspace{-.1in}
    \caption{Consider a network $G=(V,E)$ with urban travel times $\tau_{ij}=10$ minutes and airport routes $\tau_{iG}=\{19,\ldots,22\}$ minutes. Schedule three vehicles $k_i$ (capacity $c_k=2$) to deliver four passengers $r_i$ to airport during $[8:45,9:00]$.}
    \label{fig:URS-DirectedGraph}
\end{figure}

\subsubsection{URS Results}

Figure~\ref{fig:ALASmonte-carlo} shows an optimal solution with a total travel distance of 87 km, outperforming both GPT-4o-Task and DeepSeek R1 (Figure~\ref{fig:GPTDS-schedule}) which requires 123 km, a marked improvement of 41.37\%. 

\paragraph{Extended Materials}

We also used LangGraph to implement both \textbf{P3} and \textbf{P4}, as presented in \textbf{Appendix}~\ref{app:p3p4}. Furthermore, we provide a solution to \textbf{P11} for the prediction of the stock value in Code and Dataset link. These sample implementations validate the completeness of the problem statements in this benchmark.

\vspace{-.15in}
\begin{figure}[ht!]
    \centering
    \begin{tikzpicture}[
        location/.style={
            circle,
            draw=black,
            fill=white,
            minimum size=1.0cm,
            text centered,
            font=\normalsize
        },
        airport/.style={
            rectangle,
            rounded corners,
            draw=black,
            fill=cyan!20,
            minimum width=2.0cm,
            minimum height=1.0cm,
            text centered,
            font=\normalsize
        },
        passenger/.style={
            font=\small\color{red}
        },
        path_k1/.style={
            thick,
            blue,
            ->,
            >=stealth
        },
        path_k2/.style={
            thick,
            orange,
            ->,
            >=stealth
        },
        vehicle/.style={
            draw,
            fill=white,
            minimum width=0.8cm,
            minimum height=0.4cm,
            rounded corners=1pt
        }
    ]
        \node[blue] at (-1,1.5) {\Huge\faTaxi};
        \node[blue] at (-1,2.0) {k1};

        \node[location] (A) at (0,1.5) {A};
        \node[location] (B) at (2.1,1.5) {B};
        \node[location] (C) at (0,0) {C};
        \node[location] (D) at (2.1,0) {D};
        
        \node[orange] at (-1,0) {\Huge\faTaxi};
        \node[orange] at (-1,0.5) {k2};
        
        \node[airport] (G) at (5,.75) {G (BOS)};
        
        \draw[path_k1] (A) -- (B) node[midway,above,sloped,fill=none,font=\footnotesize] {8:00-8:10};
        \draw[path_k1] (B) -- (G) node[midway,above,sloped,fill=none,font=\footnotesize] {8:10-8:30};
        
        \draw[path_k2] (C) -- (D) node[midway,below,fill=none,font=\footnotesize] {8:00-8:10};
        \draw[path_k2] (D) -- (G) node[midway,below,sloped,fill=none,font=\footnotesize] {8:10-8:30};
        
        \node[passenger] at ($(A)+(0.3,0.7)$) {$r_1$ (8:45)};
        \node[passenger] at ($(B)+(0.3,0.7)$) {$r_2$ (8:50)};
        \node[passenger] at ($(C)+(0.3,0.7)$) {$r_3$ (8:55)};
        \node[passenger] at ($(D)+(0.3,0.7)$) {$r_4$ (9:00)};
        
    \end{tikzpicture}
    \caption{This solution routes showing optimal vehicle assignments. Vehicle k1 (blue) starts at A and collects passengers r1 and r2, while k2 (orange) starts at C and serves r3 and r4. Both vehicles arrive at BOS at 8:30, meeting all passenger arrival deadlines. Total distance traveled = 87 km ({\color{red}Optimal}).}
    \label{fig:ALASmonte-carlo}
    \vspace{-.1in}
\end{figure}
\vspace{-.1in}
\begin{figure}[ht!]
    \centering
    \begin{tikzpicture}[
        location/.style={
            circle,
            draw=black,
            fill=white,
            minimum size=1.0cm,
            text centered,
            font=\normalsize
        },
        airport/.style={
            rectangle,
            rounded corners,
            draw=black,
            fill=cyan!20,
            minimum width=2cm,
            minimum height=1cm,
            text centered,
            font=\normalsize
        },
        passenger/.style={
            font=\small\color{red}
        },
        path_k1/.style={
            thick,
            blue,
            ->,
            >=stealth
        },
        path_k2/.style={
            thick,
            orange,
            ->,
            >=stealth
        },
        path_k3/.style={
            thick,
            green!50!black,
            ->,
            >=stealth
        },
        vehicle/.style={
            draw,
            fill=white,
            minimum width=0.8cm,
            minimum height=0.4cm,
            rounded corners=1pt
        }
    ]


        \node[blue] at (-1,1.5) {\Huge\faTaxi};
        \node[blue] at (-1,2.0) {k1};

        \node[location] (A) at (0,1.6) {A};
        \node[location] (B) at (2.1,1.6) {B};
        \node[location] (C) at (3.8,0) {C};
        \node[location] (E) at (0,0) {E};
        \node[location] (D) at (2.0,0) {D};
        
        \node[orange] at (4.8,0) {\Huge\faTaxi};
        \node[orange] at (4.8,0.5) {k2};

        \node[green!50!black] at (-1,0) {\Huge\faTaxi};
        \node[green!50!black] at (-1,0.5) {k3};
        
        \node[airport] (G) at (5.0,1.6) {G (BOS)};
        
        \draw[path_k1] (A) -- (B) node[midway,above,font=\footnotesize] {7:40-7:50};
        \draw[path_k1] (B) -- (G) node[midway,above,fill=none,font=\footnotesize] {7:50-8:10};
        
        \draw[path_k2] (C) -- (G) node[pos=0.58,sloped,below,fill=none,font=\footnotesize] {7:55-8:17};

        \draw[path_k3] (E) -- (D) node[midway,sloped,above,fill=none,font=\footnotesize] {7:50-8:00};

        \draw[path_k3] (D) -- (G) node[pos=0.66,sloped,above,fill=none,font=\footnotesize] {8:00-8:19};
        
        \node[passenger] at ($(A)+(0.3,0.7)$) {$r_1$ (8:45)};
        \node[passenger] at ($(B)+(0.3,0.7)$) {$r_2$ (8:50)};
        \node[passenger, fill=none, inner sep=2pt] at ($(C)+(0.0,0.65)$) {$r_3$};

        \node[passenger] at ($(D)+(0.3,0.7)$) {$r_4$ (9:00)};
        
    \end{tikzpicture}
    \caption{GPT-4o-Task and DeepSeek R1 both schedule three routes. Vehicle k1 picks up r1 from A at 7:40, then r2 from B at 7:50. Vehicle k2 picks up r3 from C at 7:55. Vehicle k3 must first drive from E to D to pick up r4 at 8:00. All meet deadlines. Total travel distance = 123 km.}
    \label{fig:GPTDS-schedule}
    \vspace{-.1in}
\end{figure}

\newpage
\subsection{URS Sample Implementation}
\label{app:p3p4}

This appendix presents an implementation of \textbf{P3} and \textbf{P4}: Urban Ride Sharing without and with interrupts, using LangGraph~\cite{langgraph2024}.

\subsubsection{Agentic Workflow Formulation}

In this first stage, we define agents to manage the nodes of the workflow, including data collection, route planning, vehicle dispatch, traffic adjustment, monitoring and alert, and logging agents. At the end, we use the \texttt{>>} syntax to specify dependencies among agents. The transition from problem specifications to workflow formulation is handled automatically by ALAS\,\cite{Edward2025ALAS} in LangGraph. The code also support AutoGen, CrewAI, Swarm, and is extendable to other multi-agent framework.

\begin{lstlisting}[style=PythonStyle, caption=Collaborative Agents and Prompts, label=lst:agent_pipeline1]
    # ---- Data Collection Agent ---- #
    DC_Agent = Agent(
        name=``Data Collection Agent'',
        backstory=``You collect basic traffic data, road closure updates, and estimated travel times between locations in Bay Area suburb.'',
        task_description=``Retrieve traffic conditions, road closures, and estimated travel durations for all routes involved in passenger transport.'',
        task_expected_output=``Structured travel time data, including:
        1) CityMap: A graph G = (V, E), where the locations V and roads E have distances and travel times.
        2) Ride Requests: A set of requests R, each defined by: PassengerID, pickup/drop-off locations, and time windows.
        3) Vehicles: A set of available vehicles K, each with location, battery/fuel level, passenger capacity, and speed.'')
        
    # ---- Route Planning Agent ---- #
    RP_Agent = Agent(
        name=``Route Planning Agent'',
        backstory=``You determine optimal routes for vehicles to minimize total travel time while ensuring all passengers arrive on time.'',
        task_description=``Use traffic data and constraints to compute the best routes for each vehicle, ensuring on-time airport arrivals.'',
        task_expected_output=``Optimized vehicle assignments and travel routes.'')
    
    # ---- Vehicle Dispatch Agent ---- #
    VD_Agent = Agent(
        name=``Vehicle Dispatch Agent'',
        backstory=``You assign passengers to vehicles and ensure each vehicle follows the optimal planned route,'',
        
        task_description=``Assign passengers to vehicles based on capacity constraints and route efficiency,''
        
        task_expected_output=``Vehicle assignment list and dispatch schedule.'')
    
    # ---- Traffic Adjustment Agent (e.g. disruptions, special cases)---- #
    TA_Agent = Agent(
        name=``Traffic Adjustment Agent'',
        backstory=``You monitor live traffic updates and adjust vehicle routes dynamically in case of delays in Bay Area suburb.'',
        task_description=``Recompute routes in real time when disruptions occur (traffic, road closures), ensuring minimal delays.'',
        task_expected_output=``Updated travel plans for affected vehicles.'')
    
    # ---- Monitoring & Alert Agent ---- #
    MA_Agent = Agent(
        name=``Monitoring & Alert Agent'',
        backstory=``You track vehicle movements and notify if there are risks of missing passenger deadlines in Bay Area suburb.'',
        task_description=``Send alerts for potential delays and recommend contingency plans.'',
        task_expected_output=``Timely notifications for alternative route adjustments or emergency responses.'')
    
    Writer_agent = Agent(
        name=``Writer Agent'',
        backstory=``You are a language model specialized in writing text into .json files'',
        task_description=``Write the json response into ./p3_output.json'',
        task_expected_output=``A .json file containing the given string,
        tools=write_str_to_txt'')

    # ---- Define Dependencies (With Disruption) ---- #
    DC_Agent >> RP_Agent >> VD_Agent >> TA_Agent >> MA_Agent >> Writer_agent

    # ---- Define Dependencies (Without Disruption) ---- #
    DC_Agent >> RP_Agent >> VD_Agent >> MA_Agent >> Writer_agent

\end{lstlisting}

\subsubsection{Meta Plan Execution without Disruption (P3 URS-static)}

Now that the meta-plan workflow has been constructed, the second step involves providing real data for workflow execution. The following code snippet illustrates how vehicle and passenger locations are specified, followed by the corresponding agent executions.

\begin{lstlisting}[style=JSONStyle, caption=Sample JSON Output from Agentic Workflow, label=lst:data_collection]
==================================================
RUNNING AGENT: Data Collection Agent
==================================================
1) \textbf{CityMap: A graph G=(V,E)}
   - \textbf{Locations V:}
     - V1: Downtown
     - V2: Golden Gate Park
     - V3: Fisherman Wharf
     - V4: Union Square
     - V5: San Francisco Airport (SFO)

   - \textbf{Roads E with distances and travel times:}
     - E1: Downtown to Golden Gate Park, Distance: 5 miles, Travel Time: 20 minutes
     - E2: Golden Gate Park to Fisherman Wharf, Distance: 6 miles, Travel Time: 25 minutes
     - E3: Fisherman Wharf to Union Square, Distance: 2 miles, Travel Time: 10 minutes
     - E4: Union Square to SFO, Distance: 15 miles, Travel Time: 30 minutes
     - E5: Downtown to SFO, Distance: 13 miles, Travel Time: 35 minutes

2) \textbf{Ride Requests: A set of requests R}
   - Request 1:
     - PassengerID: P001
     - Pickup: Downtown
     - Drop-off: Union Square
     - Time Window: 9:00 AM - 9:30 AM
   - Request 2:
     - PassengerID: P002
     - Pickup: Golden Gate Park
     - Drop-off: SFO
     - Time Window: 10:00 AM - 11:00 AM
   - Request 3:
     - PassengerID: P003
     - Pickup: Fisherman Wharf
     - Drop-off: Downtown
     - Time Window: 1:00 PM - 1:30 PM

3) \textbf{Vehicles: A set of available vehicles K}
   - Vehicle 1:
     - Location: Downtown
     - Battery/Fuel Level: 80%
     - Passenger Capacity: 4
     - Speed: 30 mph
   - Vehicle 2:
     - Location: Union Square
     - Battery/Fuel Level: 60%
     - Passenger Capacity: 6
     - Speed: 25 mph
   - Vehicle 3:
     - Location: SFO
     - Battery/Fuel Level: 90%
     - Passenger Capacity: 3
     - Speed: 35 mph

This structured data provides an overview of the current traffic conditions, road closures, and estimated travel times for passenger transport in SF.

================================================== 
RUNNING AGENT: Route Planning Agent
==================================================
Optimized vehicle assignments and travel routes:

1) \textbf{Vehicle Assignments:}

   - Vehicle 1:
     - Assigned to Request 1
     - Route: Downtown (V1) -> Union Square (V4)
     - Estimated Travel Time: 20 minutes (E1 + E3)
     - Departure Time: 9:00 AM
     - Arrival Time: 9:20 AM

   - Vehicle 2:
     - Assigned to Request 2
     - Route: Union Square (V4) -> Golden Gate Park (V2) -> SFO (V5)
     - Estimated Travel Time: 55 minutes (E3 + E2 + E4)
     - Departure Time: 10:00 AM
     - Arrival Time: 10:55 AM

   - Vehicle 3:
     - Assigned to Request 3
     - Route: Fisherman Wharf (V3) -> Downtown (V1)
     - Estimated Travel Time: 10 minutes (E3)
     - Departure Time: 1:00 PM
     - Arrival Time: 1:10 PM

2) \textbf{Travel Routes:}

   - Request 1:
     - Vehicle 1 will pick up Passenger P001 at Downtown and drop off at Union Square.
     - Route: V1 -> V4

   - Request 2:
     - Vehicle 2 will pick up Passenger P002 at Golden Gate Park and drop off at SFO.
     - Route: V4 -> V2 -> V5

   - Request 3:
     - Vehicle 3 will pick up Passenger P003 at Fisherman Wharf and drop off at Downtown.

==================================================
RUNNING AGENT: Vehicle Dispatch Agent
==================================================
Vehicle Assignment List and Dispatch Schedule:

1) \textbf{Vehicle Assignments:}

   - Vehicle 1:
     - Assigned to Request 1
     - Passengers: P001
     - Route: Downtown (V1) -> Union Square (V4)
     - Estimated Travel Time: 20 minutes
     - Departure Time: ``9:00 AM''
     - Arrival Time: ``9:20 AM''

   - Vehicle 2:
     - Assigned to Request 2
     - Passengers: P002
     - Route: Union Square (V4) -> Golden Gate Park (V2) -> SFO (V5)
     - Estimated Travel Time: 55 minutes
     - Departure Time: ``10:00 AM''
     - Arrival Time: ``10:55 AM''

   - Vehicle 3:
     - Assigned to Request 3
     - Passengers: P003
     - Route: Fisherman Wharf (V3) -> Downtown (V1)
     - Estimated Travel Time: 10 minutes
     - Departure Time: ``1:00 PM''
     - Arrival Time: ``1:10 PM''

2) \textbf{Dispatch Schedule:}

   - 9:00 AM: Vehicle 1 departs from Downtown with Passenger P001.
   - 9:20 AM: Vehicle 1 arrives at Union Square.
   - 10:00 AM: Vehicle 2 departs from Union Square with Passenger P002.
   - 10:55 AM: Vehicle 2 arrives at SFO.
   - 1:00 PM: Vehicle 3 departs from Fisherman Wharf with Passenger P003.
   - 1:10 PM: Vehicle 3 arrives at Downtown.

These assignments and schedules ensure efficient use of vehicles and timely arrival of all passengers.

==================================================
RUNNING AGENT: Monitoring & Alert Agent
==================================================
\textbf{Alert Notification:}

1) Vehicle 1:
   - Current Status: On schedule.
   - Recommendation: No action needed as the vehicle is expected to arrive on time.

2) Vehicle 2:
   - Current Status: Potential delay risk due to high traffic expected around Golden Gate Park.
   - Recommendation: Consider adjusting the route to avoid traffic congestion. Possible alternative: Take the route via Sunset Blvd to bypass heavy traffic areas. Notify the passenger of the potential delay and the alternative route.

3) Vehicle 3:
   - Current Status: On schedule.
   - Recommendation: No action needed as the vehicle is expected to arrive on time.

\textbf{Contingency Plan:}

Emergency Response: If Vehicle 2 faces unexpected delays despite the alternative route, prepare a standby vehicle for immediate dispatch from a nearby location to ensure Passenger P002 reaches SFO on time.

==================================================
RUNNING AGENT: Writer Agent
==================================================
Thought: I need to write the given context string into a JSON file named ``./p3_output.json''.

Using Tool: write_str_to_txt
Tool call dict: {omitted}
Data successfully written to ``./p3_output.json''
Tool result: None
Observations: {0: None}
The context string has been successfully written into ``./p3_output.json'' as a JSON file.

\end{lstlisting}

\subsubsection{Meta Plan Execution with Disruption (P4 URS-dynamic)}

This section depicts the execution of the meta-plan (workflow) under traffic disruption. In this scenario, the system dynamically identifies traffic delays and integrates real-time updates into the planning process. Each agent plays a crucial role: the data collection agent continuously monitors traffic conditions; the route planning agent adjusts travel routes based on congestion data; and the dispatch agent ensures that vehicle assignments are optimized despite delays. The workflow demonstrates how coordinated agent interventions, including dynamic rerouting and schedule adjustments, effectively mitigate the impact of heavy traffic, maintaining timely transportation, and enhancing overall operational efficiency.

\begin{lstlisting}[style=JSONStyle, caption=Sample JSON Output from Agent Pipeline, label=lst:data_collection2]
==================================================
RUNNING AGENT: Traffic Adjustment Agent
==================================================
1) \textbf{Vehicle V1 (Passenger 101):}
   - \textbf{Current Disruption:} Heavy traffic reported on the route from Downtown SF to SF Airport.
   - \textbf{Updated Route:}
     - Start at Union Square
     - Travel to Downtown SF (2 miles, 15 minutes)
     - Detour via I-280 South instead of the usual route to SF Airport
     - Estimated travel time via detour: 30 minutes
   - \textbf{Total Updated Estimated Travel Time:} 45 minutes
   - \textbf{Revised Departure Time:} ``08:15'' to ensure timely arrival within the window.
   - \textbf{Special Instructions:} Maintain communication with the passenger about the updated route and expected arrival time.
2) \textbf{Vehicle V2 (Passenger 102):}
   - \textbf{Current Disruption:} Road closure near Fisherman Wharf due to an event.
   - \textbf{Updated Route:}
     - Start at Fisherman Wharf
     - Detour via Bay Street to Van Ness Avenue, then proceed to Golden Gate Bridge
     - Estimated travel time via detour: 15 minutes
   - \textbf{Total Updated Estimated Travel Time:} 15 minutes
   - \textbf{Revised Departure Time:} ``09:15'' to ensure timely arrival within the window.
   - \textbf{Special Instructions:} Ensure fuel levels are adequate for the detour and communicate any changes to the passenger.

\end{lstlisting}

\end{document}